\documentclass{article}

\usepackage{microtype}
\usepackage{graphicx}
\usepackage{subfigure}
\usepackage{booktabs} 
\usepackage[utf8]{inputenc}
\usepackage[T1]{fontenc}
\usepackage{hyperref}
\usepackage{url}
\usepackage{booktabs}
\usepackage{amsfonts}
\usepackage{dsfont}
\usepackage{nicefrac}
\usepackage{microtype}
\usepackage{amsmath}
\usepackage{amsthm}
\usepackage{amssymb}
\usepackage{amsfonts}
\usepackage{graphicx}
\usepackage{subfigure} 
\usepackage{stmaryrd}
\newtheorem{theorem}{Theorem}
\newtheorem{proposition}[theorem]{Proposition}
\newtheorem{definition}[theorem]{Definition}
\newtheorem{lemma}[theorem]{Lemma}

\numberwithin{theorem}{section}

\newcommand\ex{\mathbb{E}}
\newcommand\comp{\ensuremath\mathop{\scalebox{.6}{$\circ$}}}
\renewcommand\mathbb{\mathds}
\newcommand\err{\mathrm{err}}
\newcommand\disp{\mathrm{disp}}

\usepackage[accepted]{icml2019}

\icmltitlerunning{Bridging Theory and Algorithm for Domain Adaptation}

\begin{document}

\twocolumn[

	\icmltitle{Bridging Theory and Algorithm for Domain Adaptation}

	\icmlsetsymbol{equal}{*}

	\begin{icmlauthorlist}
		\icmlauthor{Yuchen Zhang}{equal,ss,lab}
		\icmlauthor{Tianle Liu}{equal,lab,math}
		\icmlauthor{Mingsheng Long}{ss,lab}
		\icmlauthor{Michael I. Jordan}{ucb}
	\end{icmlauthorlist}

	\icmlaffiliation{ss}{School of Software}
	\icmlaffiliation{math}{Department of Mathematical Science, Tsinghua University, China}
	\icmlaffiliation{lab}{Research Center for Big Data, BNRist}
	\icmlaffiliation{ucb}{University of California, Berkeley, USA.
	
	$^\dag$Yuchen Zhang <zhangyuc17@mails.tsinghua.edu.cn>}
	\icmlcorrespondingauthor{Mingsheng Long}{mingsheng@tsinghua.edu.cn}
	\icmlkeywords{Machine Learning, ICML}

	\vskip 0.3in
]

\printAffiliationsAndNotice{\icmlEqualContribution}

\begin{abstract}
	This paper addresses the problem of unsupervised domain adaption from theoretical and algorithmic perspectives. Existing domain adaptation theories naturally imply minimax optimization algorithms, which connect well with the domain adaptation methods based on adversarial learning. However, several disconnections still exist and form the gap between theory and algorithm. We extend previous theories \cite{thy:mohri2009dd,thy:shai10ad} to multiclass classification in domain adaptation, where classifiers based on the scoring functions and margin loss are standard choices in algorithm design. We introduce Margin Disparity Discrepancy, a novel measurement with rigorous generalization bounds, tailored to the distribution comparison with the asymmetric margin loss, and to the minimax optimization for easier training. Our theory can be seamlessly transformed into an adversarial learning algorithm for domain adaptation, successfully bridging the gap between theory and algorithm. A series of empirical studies show that our algorithm achieves the state of the art accuracies on challenging domain adaptation tasks.

\end{abstract}

\section{Introduction}

It is commonly assumed in learning theories that training and test data are drawn from identical distribution. If the source domain where we train a supervised learner, is substantially dissimilar to the target domain where the learner is applied, there are no possibilities for good generalization. However, we may expect to train a model by leveraging labeled data from similar yet distinct domains, which is the key machine learning setting that domain adaptation deals with \cite{cite:Book09DSS,cite:TKDE10TLSurvey}.

Remarkable theoretical advances have been achieved in domain adaptation.  \citet{thy:mohri2009dd,thy:shai10ad} provided rigorous learning bounds for unsupervised domain adaptation, a most challenging scenario in this field. These earliest theories have later been extended in many ways, from loss functions to Bayesian settings and regression problems \cite{thy:mohri2012ydsic, thy:germain2013bayes,thy:cortes2015gd}. In addition, theories based on weighted combination of hypotheses have also been developed for multiple source domain adaptation \cite{thy:crammer2008lfms, thy:mansour2008msda, mansour2009multiple, thy:judy18ms}.

On par with the theoretical findings, there are rich advances in domain adaptation algorithms. Previous work explored various techniques for statistics matching \cite{cite:TNN11TCA, cite:Arxiv14DDC, cite:ICML15DAN, cite:ICML17JAN} and discrepancy minimization \cite{cite:ICML15RevGrad,cite:JMLR16RevGrad}. Among them, adversarial learning methods come with relatively strong theoretical insights. Inspired by \citet{cite:NIPS14GAN}, these methods are built upon the two-player game between the domain discriminator and feature extractor. Current works explored adversarial learning in diverse ways, yielding state of the art results on many tasks \cite{cite:CVPR17ADDA, cite:CVPR18MCD, cite:NIPS18CDAN}.

While many domain adaptation algorithms can be roughly interpreted as minimizing the distribution discrepancy in theories, several disconnections still form non-negligible gaps between the theories and algorithms.
Firstly, domain adaptation algorithms using scoring functions lack theoretical guarantees since previous works simply studied the 0-1 loss for classification in this setting. Meanwhile, there is a gap between the widely-used divergences in theories and algorithms \cite{cite:ICML15RevGrad,cite:JMLR12MMD,cite:ICML15DAN,cite:NIPS17JDOT}.

This work aims to bridge the gaps between the theories and algorithms for domain adaptation. We present a novel theoretical analysis of classification task in domain adaptation towards explicit guidance for algorithm design. We extend existing theories to classifiers based on the scoring functions and margin loss, which is closer to the choices for real tasks. We define a new divergence, Margin Disparity Discrepancy, and provide  margin-aware generalization bounds based on Rademacher complexity, revealing that there is a trade-off between generalization error and the choice of margin. Our theory can be seamlessly transformed into an adversarial learning algorithm for domain adaptation, which achieves state of the art accuracies on several challenging real tasks.

\section{Preliminaries}

In this section we introduce basic notations and assumptions for classification problems in domain adaptation.

\subsection{Learning Setup}

In supervised learning setting, the learner receives a sample of \(n\) labeled points \(\{(x_i,y_i)\}_{i=1}^n\) from \(\mathcal X \times \mathcal Y\), where \(\mathcal X\) is an input space and \(\mathcal Y\) is an output space, which is \(\{0,1\}\) in binary classification and \(\{1,\ldots,k\}\) in multiclass classification. The sample is denoted by \(\widehat D\) if independently drawn according to the distribution \(D\).

In unsupervised domain adaptation, there are two different distributions, the source \(P\) and the target $Q$. The learner is trained on a labeled sample $\widehat P = \{(x_i^s, y_i^s)\}_{i=1}^{n}$ drawn from the source distribution and an unlabeled sample $\widehat Q = \{x_i^t\}_{i=1}^{m}$ drawn from the target distribution.

Following the notations of \citet{thy:mohri2012foundations}, we consider multiclass classification with hypothesis space \(\mathcal F\) of \emph{scoring functions} $f: \mathcal X\to \mathbb R^{| \mathcal Y|}=\mathbb R^k$, where the outputs on each dimension indicate the confidence of prediction. With a little abuse of notations, we consider \(f:\mathcal{X}\times\mathcal{Y}\to \mathbb{R}\) instead and \(f(x,y)\) indicates the component of \(f(x)\) corresponding to the label \(y\). The predicted label associated to point $x$ is the one resulting in the largest score. Thus it induces a labeling function space \(\mathcal H\) containing \(h_f\) from $\mathcal X$ to $\mathcal Y$:
\begin{equation}\label{eq:h_f}
	h_f : x\mapsto\mathop{\arg\max}_{y\in\mathcal Y}f(x,y).
\end{equation}
The \textit{(expected) error rate} and \textit{empirical error rate} of a classifier $h\in \mathcal H$ with respect to distribution $D$ are given by
\begin{equation}
	\begin{aligned}
		{\err}_D(h)            & \triangleq\ex_{(x,y)\sim D}\mathbb 1  [h(x)\neq y],         \\
		{\err}_{\widehat D}(h) & \triangleq\ex_{(x,y)\sim \widehat D}\mathbb 1  [h(x)\neq y] \\
		                       & =\frac 1 n \sum_{i=1}^n \mathbb 1  [h(x_i)\neq y_i],
	\end{aligned}
\end{equation}
where $\mathbb 1 $ is the indicator function.

Before further discussion, we assume the constant classifier $1 \in\mathcal H$ and $\mathcal H$ is closed under permutations of $\mathcal Y$. For binary classification, this is equivalent to the assumption that for any $  h\in \mathcal H$, we have $1 -h\in\mathcal H$.

\subsection{Margin Loss}

In practice, the margin between data points and the classification surface plays a significant role in achieving strong generalization performance. Thus a margin theory for classification was developed by \citet{thy:koltchinskii2002empirical}, where the 0-1 loss is replaced by the margin loss.

Define the \textit{margin} of a hypothesis $f$ at a labeled example $(x,y)$ as
\begin{equation}
	\rho_f(x,y)\triangleq\frac 1 2 (f(x,y)-\max_{y'\neq y}f(x,y')).
\end{equation}
The corresponding \textit{margin loss} and \textit{empirical margin loss} of a hypothesis $f$ is
\begin{equation}
	\begin{gathered}
		\err_{D}^{(\rho)}(f)\triangleq\ex_{x \sim D}\Phi_{\rho}\comp\rho_f(x,y),\\
		\err_{\widehat D}^{(\rho)}(f) \triangleq \ex_{x \sim \widehat D}\Phi_{\rho}\comp\rho_f(x,y)=\frac 1 n\sum_{i=1}^n \Phi_{\rho}(\rho_f(x_i,y_i)),
	\end{gathered}
\end{equation}
where $\comp$ denotes function composition, and $\Phi_\rho$ is
\begin{equation}
	\Phi_\rho(x)\triangleq\begin{cases}0 & \rho\leq x\\1-x/\rho& 0\leq x\leq \rho\\1& x\leq 0\end{cases}.
\end{equation}
An important property is that $\err_D^{(\rho)}(f)\geq \err_D(h_f)$ for any $\rho>0$ and $f\in \mathcal F$. \citet{thy:koltchinskii2002empirical} showed that the margin loss leads to an informative generalization bound for classification. Based on this seminal work, we shall develop \emph{margin} bounds for classification in domain adaptation.

\section{Theoretical Guarantees}
\label{sec:theory}

In this section, we give theoretical guarantees for domain adaptation. \textbf{\textit{All proofs can be found in Appendices A--C.}}

To reduce the error rate on target domain with labeled training data only on source domain, the distributions \(P\) and \(Q\) should not be dissimilar substantially. Thus a measurement of their discrepancy is crucial in domain adaptation theory.

In the seminal work \cite{thy:shai10ad}, the {$\mathcal H\Delta\mathcal H$-divergence} was proposed to measure such discrepancy,
\begin{equation}
	d_{\mathcal H\Delta\mathcal H} = \sup_{h, h'\in \mathcal H}
	\left|\mathbb{E}_{ Q} \mathbb{1}[h'\ne h] - \mathbb{E}_{ P} \mathbb{1}[h'\ne h] \right|.
\end{equation}
\citet{thy:mohri2009dd} extended the $\mathcal H\Delta\mathcal H$-divergence to general loss functions, leading to the {discrepancy distance}:
\begin{equation}
	\mathrm{disc}_{L} = \sup_{h, h'\in \mathcal H}
	|\mathbb{E}_{ Q} L(h',h) - \mathbb{E}_{ P}L(h',h)|,
\end{equation}
where $L$ should be a bounded function satisfying symmetry and triangle inequality. Note that many widely-used losses, e.g. margin loss, do not satisfy these requirements.

With these discrepancy measures, generalization bounds based on VC-dimension and Rademacher complexity were rigorously derived for domain adaptation. While these theories have made influential impact in advancing algorithm designs, there are two crucial directions for improvement:
\begin{enumerate}
	\item Generalization bound for classification with \emph{scoring functions} has not been formally studied in the domain adaptation setting. As scoring functions with margin loss provide informative generalization bound in the standard classification, there is a strong motivation to develop a \emph{margin theory} for domain adaptation.
	\item The hypothesis-induced discrepancies require taking supremum over hypothesis space $\mathcal H \Delta \mathcal H$, while achieving lower generalization bound requires minimizing these discrepancies adversarially. Computing the supremum requires an ergodicity over \(\mathcal{ H}\Delta \mathcal{H}\) and the optimal hypotheses in this problem might differ significantly from the optimal classifier, which highly increases the difficulty of optimization. Thus there is a critical need for theoretically justified algorithms which minimize not only the empirical error on the source domain, but also the discrepancy measure.
\end{enumerate}

These directions are the pain points in practical algorithm designs. While designing a domain adaptation algorithm using scoring functions, we may suspect whether the algorithm is theoretically guaranteed since there is a gap between the loss functions used in the theories and algorithms. Another gap lies between the hypothesis-induced discrepancies in theories and the widely-used  divergences in domain adaptation algorithms, including Jensen Shannon Divergence \cite{cite:ICML15RevGrad}, Maximum Mean Discrepancy \cite{cite:ICML15DAN}, and Wasserstein Distance \cite{cite:NIPS17JDOT}. In this work, we aim to bridge these gaps between the theories and algorithms for domain adaptation, by defining a novel, theoretically-justified margin disparity discrepancy.

\subsection{Margin Disparity Discrepancy}
First, we give an improved discrepancy for measuring the distribution difference by restricting the hypothesis space.

Given two hypotheses \(h,h'\in \mathcal H\), we define the \emph{(expected) 0-1 disparity} between them as
\begin{equation}
	\disp_D(h',h)\triangleq\ex_{D}\mathbb 1[h'\neq h],
\end{equation}
and the \emph{empirical 0-1 disparity} as
\begin{equation}
	\disp_{\widehat D}(h',h)\triangleq\!\ex_{\widehat D}\mathbb 1[h'\neq h]=\!\frac 1 n \sum_{i=1}^n \mathbb 1 [h'(x_i)\neq h(x_i)].
\end{equation}

\begin{definition}[\textbf{Disparity Discrepancy, DD}]
	Given a hypothesis space \(\mathcal H\) and a \emph{specific classifier} \(h\!\in \!\mathcal H\), the {Disparity Discrepancy} (DD) induced by \(h' \in {\mathcal{H}}\) is defined by
	\begin{equation}
		\begin{aligned}
			d_{h,\mathcal{H}}(P,Q) & \triangleq\sup_{h'\in \mathcal H}(\disp_Q(h',h)-\disp_P(h',h)) \\ & = \sup_{h'\in \mathcal{H}}(\mathbb{E}_{ Q} \mathbb{1}[h'\ne h] - \mathbb{E}_{ P} \mathbb{1}[h'\ne h] ).
		\end{aligned}
	\end{equation}
	Similarly, the {empirical disparity discrepancy is}
	\begin{equation}
		d_{h,\mathcal{H}}(\widehat P,\widehat Q) \triangleq \sup_{h'\in \mathcal H}(\disp_{\widehat Q}(h',h)-\disp_{\widehat P}(h',h)).
	\end{equation}
\end{definition}

Note that the disparity discrepancy is not only dependent on the hypothesis space $\mathcal{H}$, but also on a specific classifier $h$. We shall prove that this discrepancy can well measure the difference of distributions (actually a pseudo-metric in the binary case) and leads to a {VC-dimension} generalization bound for binary classification. An alternative analysis of this standard case is provided in Appendix  B. Compared with the \(\mathcal H\Delta \mathcal H\)-divergence, the supremum in the disparity discrepancy is taken only over the hypothesis space \(\mathcal H\) and thus can be optimized more easily. This will significantly ease the minimax optimization widely used in many domain adaptation algorithms.

In the case of multiclass classification, the \emph{margin} of scoring functions becomes an important factor for informative generalization bound, as envisioned by \citet{thy:koltchinskii2002empirical}.
Existing domain adaptation theories \cite{cite:NIPS07DAT,thy:shai10ad,thy:blitzer2007LBDA,thy:mohri2009dd} do not give a formal analysis of generalization bound with scoring functions and margin loss. To bridge the gap between theories that typically analyze labeling functions and loss functions with symmetry and subadditivity, and algorithms that widely adopt scoring functions and margin losses, we propose a margin based disparity discrepancy.

The \emph{margin disparity}, i.e., disparity by changing the 0-1 loss to the margin loss, and {its empirical version} from hypothesis \(f\) to \(f'\) are defined as
\begin{equation}
	\begin{aligned}
		\disp_D^{(\rho)}(f',f)            & \triangleq\ex_{D}\Phi_{\rho}\comp\rho_{f'}(\cdot,h_f),         \\
		\disp_{\widehat D}^{(\rho)}(f',f) & \triangleq\ex_{\widehat D}\Phi_{\rho}\comp\rho_{f'}(\cdot,h_f) \\ &=\frac 1 n \sum_{i=1}^n \Phi_\rho\comp\rho_{f'}(x_i,h_f(x_i)).\end{aligned}
\end{equation}
Note that $f$ and $f'$ are scoring functions while $h_f$ and $h_{f'}$ are their labeling functions.
Note also that the margin disparity is not a symmetric function on $f$ and $f'$, and the generalization theory w.r.t. this loss could be quite different from that for the discrepancy distance \cite{thy:mohri2009dd}, which requires symmetry and subadditivity.

\begin{definition}[\textbf{Margin Disparity Discrepancy, MDD}]
	With the definition of margin disparity, we define Margin Disparity Discrepancy (MDD) and its {empirical version} by
	\begin{equation}
		\begin{aligned}
			 & d_{f,\mathcal F}^{(\rho)}( P, Q)\triangleq\sup_{f'\in\mathcal F}\Big{(}\disp_{ Q}^{(\rho)}(f',f)-\disp_{P}^{(\rho)}(f',f)\Big{)},
			\\ &
			d_{f,\mathcal F}^{(\rho)}(\widehat P,\widehat Q)\triangleq\sup_{f'\in\mathcal F}\Big{(}\disp_{\widehat Q}^{(\rho)}(f',f)-\disp_{\widehat P}^{(\rho)}(f',f)\Big{)}.\end{aligned}
	\end{equation}
\end{definition}

The margin disparity discrepancy (MDD) is well-defined since $d_{f,\mathcal F}^{(\rho)}(P,P) = 0$ and it satisfies the nonnegativity and subadditivity. Despite of its asymmetry, MDD has the ability to measure the distribution difference in domain adaptation regarding the following proposition.

\begin{proposition}\label{the:bound1}
	For every scoring function \(f\),
	\begin{equation}
		\err_{Q} (h_f) \le \err_P^{(\rho)} (f) +  d_{f,\mathcal{F}}^{(\rho)}(P,Q) +    \lambda,
	\end{equation}
	where
	\(\lambda=\lambda(\rho,\mathcal{F},P,Q) \) is the ideal combined margin loss:
	\begin{equation}
		\lambda = \mathop{\min}\limits_{f^*\in \mathcal{H}} \{\err_P^{(\rho)} (f^*)+\err_Q^{(\rho)}(f^*)\}.
	\end{equation}
\end{proposition}

This upper bound has a similar form with the learning bound proposed by \citet{thy:shai10ad}. $ \lambda$ is determined by the learning problem quantifying the inverse of ``\emph{adaptability}'' and can be reduced to a rather small value if the hypothesis space is rich enough. $\err_P^{(\rho)} (f) $ depicts the performance of $f$ on source domain and MDD bounds the performance gap caused by domain shift. This \emph{margin bound} gives a new perspective for analyzing domain adaptation with respect to scoring functions and margin loss.

\subsection{Domain Adaptation: Generalization Bounds}

In this subsection, we provide several generalization bounds for multiclass domain adaptation based on margin loss and margin disparity discrepancy (MDD).
First, we present a Rademacher complexity bound for the difference between MDD and its empirical version.
Then, we combine the Rademacher complexity bound of MDD and Proposition~\ref{the:bound1} to derive the final generalization bound. 

To begin with, we introduce a new function class \(\Pi_{\mathcal H}\mathcal F\) that serves as a ``scoring'' version of the symmetric difference hypothesis space \(\mathcal H \Delta \mathcal H\) in \citet{thy:shai10ad}. For more intuition, we also provide a geometric interpretation of this notion in the Appendix (Definition C.3).

\begin{definition}
	Given a class of scoring functions \(\mathcal F\) and a class of the induced classifiers \(\mathcal H\), we define $\Pi_\mathcal{H}\mathcal{F}$ as
	\begin{equation}
		\Pi_{\mathcal H}\mathcal{F} = \{x \mapsto f(x, h(x)) | h\in \mathcal{H}, f\in \mathcal{F} \}.
	\end{equation}
\end{definition}

Now we introduce the Rademacher complexity, commonly used in the generalization theory as a measurement of richness for a particular hypothesis space \cite{thy:mohri2012foundations}.

\begin{definition}[\textbf{Rademacher Complexity}]\label{def:rade}
	Let $\mathcal F$ be a family of functions mapping from $\mathcal Z=\mathcal X\times\mathcal Y$ to $[a,b]$ and $\widehat D=\{z_1,\ldots,z_n\}$ a fixed sample of size $n$ drawn from the  distribution $D$ over $\mathcal Z$. Then, the {empirical Rademacher complexity} of $\mathcal F$ with respect to the sample $\widehat D$ is defined as
	\begin{equation}
		\widehat{\mathfrak{R}}_{\widehat D}(\mathcal F)\triangleq\ex_\sigma \sup_{f\in \mathcal F}\frac 1 n \sum_{i=1}^n\sigma_if(z_i).
	\end{equation}
	where $\sigma_i$'s are independent uniform random variables taking values in $\{-1,+1\}$. The {Rademacher complexity} is
	\begin{equation}
		\mathfrak R_{n, D}(\mathcal F)\triangleq\ex_{\widehat D\sim D^n}\widehat{\mathfrak{R}}_{\widehat D}(\mathcal F).
	\end{equation}
\end{definition}

With the Rademacher complexity, we proceed to show that MDD can be well estimated through finite samples.

\begin{lemma}\label{the:mdd1}
	For any $\delta > 0$, with probability $1-2\delta$, the following holds simultaneously for any scoring function \(f\),
	\begin{equation}
		\begin{aligned}
			 & |d_{f,\mathcal F}^{(\rho)}(\widehat P,\widehat Q) -d_{f,\mathcal F}^{(\rho)}(P,Q)|
			\\
			 \leq& \frac{2k}{\rho}{\mathfrak{R}}_{n, P}(\Pi_{\mathcal H}\mathcal F) +
			\frac{2k}{\rho}{\mathfrak{R}}_{m, Q}(\Pi_{\mathcal H}\mathcal F)
			+\sqrt{\frac{\log \frac{2}{\delta}}{2n}} +\sqrt{\frac{\log \frac{2}{\delta}}{2m}}.
		\end{aligned}
	\end{equation}
\end{lemma}

This lemma justifies that the expected MDD with respect to \(f\) can be uniformly approximated by the empirical one computed on samples. The error term is controlled by the complexity of hypothesis set, the margin \(\rho\), the class number \(k\) and sample sizes \(n,m\).

Combining Proposition~\ref{the:bound1} and Lemma~\ref{the:mdd1}, we obtain a Rademacher complexity based generalization bound of the expected target error through the empirical MDD.

\begin{theorem}[\textbf{Generalization Bound}]\label{the:bound2}
	Given the same settings with Definition \ref{def:rade}, for any $\delta > 0$, with probability $1-3\delta$, we have the following uniform generalization bound for all scoring functions \(f\),
	\begin{equation}
		\begin{aligned}
			\err_Q(f)\leq & \err_{\widehat P}^{(\rho)}(f)+d_{f,\mathcal F}^{(\rho)}(\widehat P,\widehat Q) +\lambda
			\\
			+             & \frac{2k^2}{\rho}{\mathfrak{R}_{n, P}}(\Pi_1\mathcal F) +\frac{2k}{\rho}{\mathfrak{R}_{n, P}}(\Pi_{\mathcal H}\mathcal F)+2\sqrt{\frac{\log \frac 2 {\delta}}{2n}}
			\\
			\ +           & \frac{2k}{\rho}{\mathfrak{R}_{m, Q}}(\Pi_{\mathcal H}\mathcal F)+\sqrt{\frac{\log \frac{2}{\delta}}{2m}},
		\end{aligned}
	\end{equation}
	where \(\Pi_1(\mathcal{F})\) is defined as
	\begin{equation}
		\Pi_1\mathcal F\triangleq\{x\mapsto f(x,y)\big{|}y\in \mathcal Y,f\in \mathcal F\},
	\end{equation}
	and
	\(\lambda=\lambda(\rho,\mathcal{F},P,Q) \) is a constant independent of \(f\).
\end{theorem}

Note that the notation \(\Pi_1 \mathcal{F}\) follows from \citet{thy:mohri2012foundations}, where \(1\) stands for constant functions mapping all points to the same class and \(\Pi_1 \mathcal F\) can be seen as the union of projections of \(\mathcal{F}\) onto each dimension (See Appendix Lemma C.4). Such projections are needed because the Rademacher complexity is only defined for real-valued function classes.

Compared with the bounds based on 0-1 loss and $\mathcal{H}\Delta\mathcal{H}$-divergence \cite{thy:shai10ad,thy:mohri2009dd}, this generalization bound is more informative. Through choosing a better margin $\rho$, we could achieve better generalization ability on the target domain. Moreover, we point out that there is a trade-off between generalization and optimization in the choice of \(\rho\). For relatively small \(\rho\) and rich hypothesis space, the first two terms do not differ too much according to \(\rho\) so the right-hand side becomes smaller with the increase of \(\rho\). However, for too large \(\rho\), these terms cannot be optimized to reach an acceptable small value.

Although we have shown the margin bound, the value of the Rademacher complexity in Theorem~\ref{the:bound2} is still not explicit enough. Therefore, we include an example of linear classifiers in the Appendix (Example C.9).
Also we need to check the variation of ${\mathfrak{R}_{n, D}}(\Pi_{\mathcal H}\mathcal F)$ with the growth of $n$. To this end, we describe the notion of covering number from \citet{thy:zhou2002covering, thy:anthony2009neural,thy:talagrand2014upper}.

Intuitively a \emph{covering number} \(\mathcal N_2(\tau, \mathcal G)\) is the minimal number of \(\mathcal L_2\) balls of radius \(\tau>0\) needed to cover a class \(\mathcal G\) of bounded functions \(g:\mathcal X\to \mathbb R\) and can be interpreted as a measure of the richness of the class \(\mathcal G\) at scale \(\tau\). A rigorous definition is given in the Appendix together with a proof of the following covering number bound for MDD.

\begin{theorem}[\textbf{Generalization Bound with Covering Number}]\label{the:bound3}
	With the same conditions in Theorem \ref{the:bound2}, further suppose \(\Pi_1\mathcal F \) is bounded in \(\mathcal L_2\) by \(L\). For \(\delta>0\), with probability \(1-3\delta\), we have the following uniform generalization bound for all scoring functions \(f\),
	\begin{equation}
		\begin{aligned}
			\err_Q & (f)\leq \err_{\widehat P}^{(\rho)}(f)+d_{f,\mathcal F}^{(\rho)}(\widehat P,\widehat Q)+\lambda+2\sqrt{\frac{\log \frac 2 \delta}{2n}}                                                                                     \\
			       & +\sqrt{\frac{\log \frac 2 \delta}{2m}}
			+\frac{16k^2\sqrt {k}}{\rho}\inf_{\epsilon\geq 0}\Big \{\epsilon+3\big(\frac 1 {\!\sqrt n}\!\!+\!\!\frac 1 {\!\sqrt m}\big)                                                                                                        \\
			       & \big(\!\!\int_{\epsilon}^L\!\!\!\!\!\!\! \sqrt{\log \mathcal N_2(\tau, \Pi_1\mathcal F)}\textrm d \tau\!+\!L\!\!\int_{\epsilon/L}^1\!\!\!\!\!\!\! \sqrt{\log \mathcal N_2(\tau,\Pi_1\mathcal H)}\mathrm d \tau\big)\Big\}.
		\end{aligned}
	\end{equation}
\end{theorem}

Compared with \ref{the:bound2}, the Rademacher complexity terms are replaced by more intuitive and concrete notions of covering numbers. Theoretically, covering numbers also serve as a bridge between Rademacher complexity of \(\Pi_{\mathcal{H}}\mathcal{F}\) and VC-dimension style bound when \(k=2\). To show this we need the notion of \textit{fat-shattering dimension} \cite{thy:mendelson2003entropy, thy:rakhlin2014statistical}. For concision, we leave the definition and results to the Appendix (Theorem C.19), where we show that our results coincide with \citet{thy:shai10ad} in the order of sample complexity.

In summary, our theory is a bold attempt towards filling the two gaps mentioned at the beginning of this section. Firstly, we provide a thorough analysis for multiclass classification in domain adaptation. Secondly, our bound is based on scoring functions and margin loss. Thirdly, as the measure of distribution shift, MDD is defined by simply taking supremum over a single hypothesis space $\mathcal{F}$, making the minimax optimization problem easier to solve.

\section{Algorithm}

According to the above theory, we propose an adversarial representation learning method for domain adaptation.

\subsection{Minimax Optimization Problem}

Recall that the expected error $\err_{Q}(f)$ on target domain is bounded by the sum of four terms: empirical margin error on the source domain $\err_{\widehat P}^{(\rho)}(f)$, empirical MDD $d_{f,\mathcal F}^{(\rho)}(\widehat P,\widehat Q)$, the ideal error $\lambda$ and complexity terms. We need to solve the following minimization problem for the optimal classifier \(f\) in hypothesis space \(\mathcal{F}\):
\begin{equation}
	\min_{f\in\mathcal F}\; \err_{\widehat P}^{(\rho)}(f) + d_{f,\mathcal F}^{(\rho)}(\widehat P,\widehat Q).
\end{equation}

Minimizing margin disparity discrepancy is a \emph{minimax game} since MDD is defined as the supremum over hypothesis space $\mathcal{F}$. Because the max-player is still too strong, we introduce a feature extractor $\psi$ to make the min-player stronger. Applying \(\psi\) to the source and target empirical distributions, the overall optimization problem can be written as
\begin{equation}
	\begin{gathered}
		\min_{f,\psi} \; \err_{\psi(\widehat P)}^{(\rho)}(f) + (\disp_{\psi(\widehat Q)}^{(\rho)}(f^*,f)-\disp_{\psi(\widehat P)}^{(\rho)}(f^*,f)),
		\\
		f^*=\max_{f'} \; (\disp_{\psi(\widehat Q)}^{(\rho)}(f',f)-\disp_{\psi(\widehat P)}^{(\rho)}(f',f)).
	\end{gathered}
\end{equation}
To enable representation-based domain adaptation, we need to learn new representation $\psi$ such that MDD is minimized. 

\begin{figure}[H]
	\centering
	\includegraphics[width=1\columnwidth]{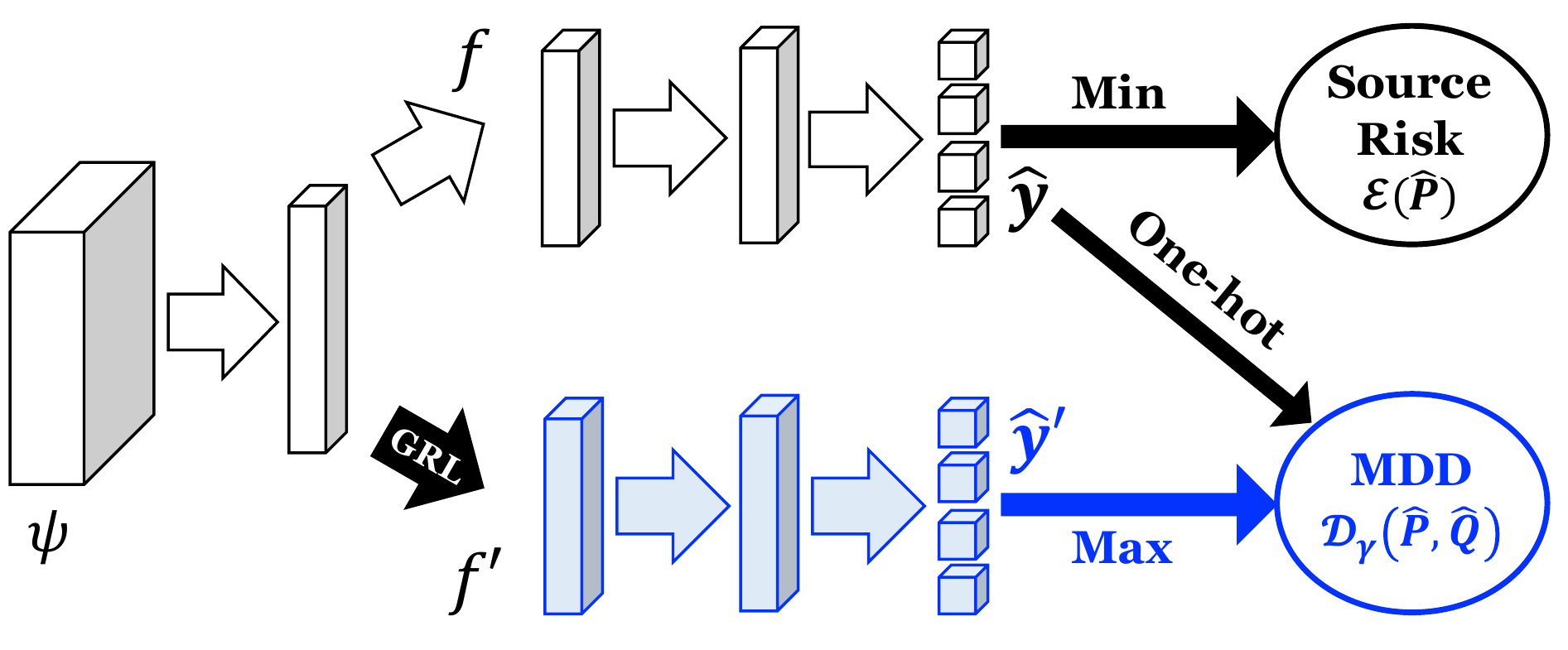}
	\vspace{-10pt}
	\caption{The adversarial network for algorithm implementation.}
	\label{fig:arch}
	\vspace{-10pt}
\end{figure}

Now we design an adversarial learning algorithm to solve this problem by introducing an auxiliary classifier $f'$ sharing the same hypothesis space with $f$. This is natively implemented in an adversarial network as Figure~\ref{fig:arch}. Also since the margin loss is hard to optimize via stochastic gradient descent (SGD) in practice, we use a combination of loss functions \(L\) and \(L'\) in substitution to the margin loss, which well preserve the key property of the margin. The practical optimization problem in the adversarial learning is stated as
\begin{equation}
	\begin{gathered}
		\min_{f,\psi} \; \mathcal{E}(\widehat P) + \eta \mathcal{D}_{\gamma}(\widehat{P}, \widehat{Q}),
		\\
		\max_{f'} \; \mathcal{D}_{\gamma}(\widehat{P}, \widehat{Q}),
	\end{gathered}
\end{equation}
where \(\eta\) is the trade-off coefficient between source error $\mathcal{E}(\widehat P)$ and MDD $\mathcal{D}_{\gamma}(\widehat{P}, \widehat{Q})$, \(\gamma \triangleq \exp \rho\) is designed to attain the margin \(\rho\) (detailed in the next subsection). Concretely,
\begin{equation}
	\begin{aligned}
		\mathcal{E}(\widehat P) & = \mathbb{E}_{(x^s , y^s)\sim \widehat{P} } \, L(f(\psi(x^s)), y^s), \\
		\mathcal{D}_\gamma(\widehat{P}, \widehat{Q})
		                        & =\mathbb E_{x^t \sim \widehat Q}L'(f'(\psi(x^t)),f(\psi(x^t)))       \\
		                        & -\gamma\mathbb E_{x^s \sim \widehat P}L(f'(\psi(x^s)),f(\psi(x^s))).
	\end{aligned}
\end{equation}
Since the discrepancy loss term is not differentiable on the parameters of $f$, for simplicity we directly train the feature extractor $\psi$ to minimize the discrepancy loss term through a gradient reversal layer (GRL) \cite{cite:ICML15RevGrad}.

\subsection{Combined Cross-Entropy Loss}

As we mentioned above, multiclass margin loss or hinge loss causes the problem of gradient vanishing in stochastic gradient descent, and thus cannot be optimized efficiently, especially for representation learning that significantly relies on gradient propagation. To overcome this common issue, we choose different loss functions on source and target and reweigh them to approximate MDD.

Denote by \(\sigma\) the softmax function, i.e., for \(\mathbf{z}\in\mathbb R^k\)
\begin{equation}
	\sigma_{j} (\mathbf {z} )={\frac {e^{z_{j}}}{\sum _{i=1}^k e^{z_{i}}}}, \;  \mathrm{for }\ j=1,\ldots,k.
\end{equation}

On the source domain, \(\err_{\widehat P}^{(\rho)}(f)\) and \(\disp_{\widehat P}^{(\rho)}(f',f)\) are replaced by the standard cross-entropy loss
\begin{equation}
	\begin{aligned}
		L(f(\psi(x^s)),y^s)           & \triangleq -\log [\sigma_{y^s}(f(\psi(x^s)))],              \\
		L(f'(\psi(x^s)),f(\psi(x^s))) & \triangleq -\log [\sigma_{h_f(\psi(x^s))}( f'(\psi(x^s)))].
	\end{aligned}
\end{equation}

On the target domain, we use a modified cross-entropy loss
\begin{equation}
	L'(f'(\psi(x^t)),\!f(\psi(x^t))) \triangleq \log [1-\sigma_{h_f(\psi(x^t))}(f'(\psi(x^t\!)))].
\end{equation}

Note that this modification was introduced in \citet{cite:NIPS14GAN} to mitigate the burden of exploding or vanishing gradients when performing adversarial learning. Combining the above two terms with a coefficient \(\gamma\), the objective of the auxiliary classifier $f'$ can be formulated as
\begin{equation}\label{loss:gamma}
	\begin{aligned}
		\max_{f'} \; \gamma \, & \ex_{x^s\sim\widehat{P}}\log [\sigma_{h_f(\psi(x^s))}( f'(\psi(x^s)))]     \\
		+ \,                   & \ex_{x^t \sim \widehat{Q}}\log [1-\sigma_{h_f(\psi(x^t))}(f'(\psi(x^t)))].
	\end{aligned}
\end{equation}
We shall see that training the feature extractor $\psi$ to minimize loss function \eqref{loss:gamma} will lead to $\psi(\widehat P) \approx \psi (\widehat Q)$.

\begin{proposition}
	(Informal) Assuming that there is no restriction on the choice of $f'$ and $\gamma > 1$, the global minimum of the loss function (\ref{loss:gamma}) is $P=Q$. The value of $\sigma_{h_f}(f'(\cdot)) $ at equilibrium is ${\gamma}/{(1+\gamma) }$\label{exp:gamma} and the corresponding margin of $f'$ is $\log \gamma$.
\end{proposition}

We refer to $\gamma = \exp \rho$ as the margin factor, with explanation given in the Appendix (Theorems D.1 \& D.2). In general larger \(\gamma\) yields better generalization. However, as we have explained in Section~\ref{sec:theory}, we cannot let it go to infinity. In fact, from an empirical view \(\rho\) can only be chosen far beyond the theoretical optimal value since performing SGD for a large \(\gamma\) might lead to exploding gradients. In summary, the choice of \(\gamma\) is crucial in our method and we prefer relatively larger \(\gamma\) in practice when exploding gradients are not encountered.

\begin{table*}[htbp]
	\addtolength{\tabcolsep}{2pt}
	\centering
	\vspace{-5pt}
	\caption{Accuracy (\%) on {Office-31} for unsupervised domain adaptation (ResNet-50).}
	\label{table:office31}
	\vskip 0.05in
	\resizebox{\textwidth}{!}{%
		\begin{tabular}{lccccccc}
			\toprule
			Method                          & A $\rightarrow$ W     & D $\rightarrow$ W     & W $\rightarrow$ D     & A $\rightarrow$ D     & D $\rightarrow$ A     & W $\rightarrow$ A     & Avg           \\
			\midrule
			ResNet-50 \cite{cite:CVPR16DRL} & 68.4$\pm$0.2          & 96.7$\pm$0.1          & 99.3$\pm$0.1          & 68.9$\pm$0.2          & 62.5$\pm$0.3          & 60.7$\pm$0.3          & 76.1          \\
			DAN \cite{cite:ICML15DAN}       & 80.5$\pm$0.4          & 97.1$\pm$0.2          & 99.6$\pm$0.1          & 78.6$\pm$0.2          & 63.6$\pm$0.3          & 62.8$\pm$0.2          & 80.4          \\
			DANN \cite{cite:JMLR16RevGrad}  & 82.0$\pm$0.4          & 96.9$\pm$0.2          & 99.1$\pm$0.1          & 79.7$\pm$0.4          & 68.2$\pm$0.4          & 67.4$\pm$0.5          & 82.2          \\
			ADDA \cite{cite:CVPR17ADDA}     & 86.2$\pm$0.5          & 96.2$\pm$0.3          & 98.4$\pm$0.3          & 77.8$\pm$0.3          & 69.5$\pm$0.4          & 68.9$\pm$0.5          & 82.9          \\
			JAN \cite{cite:ICML17JAN}       & 85.4$\pm$0.3          & {97.4}$\pm$0.2        & {99.8}$\pm$0.2        & 84.7$\pm$0.3          & 68.6$\pm$0.3          & 70.0$\pm$0.4          & 84.3          \\
			GTA \cite{cite:CVPR18GTA}       & 89.5$\pm$0.5          & 97.9$\pm$0.3          & 99.8$\pm$0.4          & 87.7$\pm$0.5          & 72.8$\pm$0.3          & 71.4$\pm$0.4          & 86.5          \\
			MCD \cite{cite:CVPR18MCD}   &88.6$\pm$0.2&98.5$\pm$0.1&\textbf{100.0}$\pm$.0&92.2$\pm$0.2&69.5$\pm$0.1&69.7$\pm$0.3&86.5 \\
			CDAN \cite{cite:NIPS18CDAN}     & 94.1$\pm$0.1          & \textbf{98.6}$\pm$0.1 & \textbf{100.0}$\pm$.0 & 92.9$\pm$0.2          & 71.0$\pm$0.3          & 69.3$\pm$0.3          & 87.7          \\
			\textbf{MDD} (Proposed)                  & \textbf{94.5}$\pm$0.3 & 98.4$\pm$0.1          & \textbf{100.0}$\pm$.0 & \textbf{93.5}$\pm$0.2 & \textbf{74.6}$\pm$0.3 & \textbf{72.2}$\pm$0.1 & \textbf{88.9} \\
			\bottomrule
		\end{tabular}
	}
	\vspace{-10pt}
\end{table*}

\begin{table*}[htbp]
	\addtolength{\tabcolsep}{-5pt}
	\centering
	\caption{Accuracy (\%) on {Office-Home} for unsupervised domain adaptation (ResNet-50).}
	\label{table:officehome}
	\vskip 0.05in
	\resizebox{\textwidth}{!}{%
		\begin{tabular}{lccccccccccccc}
			\toprule
			Method                          & Ar$\shortrightarrow$Cl & Ar$\shortrightarrow$Pr & Ar$\shortrightarrow$Rw & Cl$\shortrightarrow$Ar & Cl$\shortrightarrow$Pr & Cl$\shortrightarrow$Rw & Pr$\shortrightarrow$Ar & Pr$\shortrightarrow$Cl & Pr$\shortrightarrow$Rw & Rw$\shortrightarrow$Ar & Rw$\shortrightarrow$Cl & Rw$\shortrightarrow$Pr & Avg           \\
			\midrule
			ResNet-50 \cite{cite:CVPR16DRL} & 34.9                   & 50.0                   & 58.0                   & 37.4                   & 41.9                   & 46.2                   & 38.5                   & 31.2                   & 60.4                   & 53.9                   & 41.2                   & 59.9                   & 46.1          \\
			DAN \cite{cite:ICML15DAN}       & 43.6                   & 57.0                   & 67.9                   & 45.8                   & 56.5                   & 60.4                   & 44.0                   & 43.6                   & 67.7                   & 63.1                   & 51.5                   & 74.3                   & 56.3          \\
			DANN \cite{cite:JMLR16RevGrad}  & 45.6                   & 59.3                   & 70.1                   & 47.0                   & 58.5                   & 60.9                   & 46.1                   & 43.7                   & 68.5                   & 63.2                   & 51.8                   & 76.8                   & 57.6          \\
			JAN \cite{cite:ICML17JAN}       & 45.9                   & 61.2                   & 68.9                   & 50.4                   & 59.7                   & 61.0                   & 45.8                   & 43.4                   & 70.3                   & 63.9                   & 52.4                   & 76.8                   & 58.3          \\
			CDAN \cite{cite:NIPS18CDAN}     & 50.7                   & 70.6                   & 76.0                   & 57.6                   & 70.0                   & 70.0                   & 57.4                   & 50.9                   & 77.3                   & 70.9                   & 56.7                   & 81.6                   & 65.8          \\
			\textbf{MDD} (Proposed)                & \textbf{54.9}          & \textbf{73.7}          & \textbf{77.8}          & \textbf{60.0}          & \textbf{71.4}          & \textbf{71.8}          & \textbf{61.2}          & \textbf{53.6}          & \textbf{78.1}          & \textbf{72.5}          & \textbf{60.2}          & \textbf{82.3}          & \textbf{68.1}
			\\
			\bottomrule
		\end{tabular}%
	}
	\vspace{-10pt}
\end{table*}

\section{Experiments}
We evaluate the proposed learning method on three datasets against state of the art deep domain adaptation methods. The code is available at \url{github.com/thuml/MDD}.

\subsection{Setup}

\textbf{Office-31} \cite{cite:ECCV10Office} is a standard domain adaptation dataset of three diverse domains, \textbf{A}mazon from Amazon website, \textbf{W}ebcam by web camera and \textbf{D}SLR by digital SLR camera with 4,652 images in 31 unbalanced classes.

\textbf{Office-Home} \cite{cite:CVPR17DHN}  is a more complex dataset containing 15,500 images from four visually very different domains: \textbf{Ar}tistic images, \textbf{Cl}ip Art, \textbf{Pr}oduct images, and \textbf{R}eal-\textbf{w}orld images.

\textbf{VisDA-2017} \cite{cite:VISDA2017} is simulation-to-real dataset with two extremely distinct domains: \textbf{Synthetic} renderings of 3D models and \textbf{Real} collected from photo-realistic or real-image datasets. With 280K images in 12 classes, the scale of VisDA-2017 brings challenges to domain adaptation.

We compare our designed algorithm based on Margin Disparity Discrepancy (\textbf{MDD}) with state of the art domain adaptation methods: Deep Adaptation Network (\textbf{DAN}) \cite{cite:ICML15DAN},  Domain Adversarial Neural Network (\textbf{DANN}) \cite{cite:JMLR16RevGrad},
Joint Adaptation Network (\textbf{JAN}) \cite{cite:ICML17JAN}, Adversarial Discriminative Domain Adaptation (\textbf{ADDA}) \cite{cite:CVPR17ADDA}, Generate to Adapt (\textbf{GTA}) \cite{cite:CVPR18GTA}, Maximum Classifier Discrepancy (\textbf{MCD}) \cite{cite:CVPR18MCD}, and Conditional Domain Adversarial Network  (\textbf{CDAN)} \cite{cite:NIPS18CDAN}.

We follow the commonly used experimental protocol for unsupervised domain adaptation from \citet{cite:ICML15RevGrad, cite:NIPS18CDAN}. We report the average accuracies of five independent experiments. The importance-weighted cross-validation (\textbf{IWCV}) is employed in all experiments for the selection of hyper-parameters. The asymptotic value of coefficient $\eta$ is fixed to $0.1$ and $\gamma$ is chosen from $\{2,3,4\}$ and kept the same for all tasks on the same dataset.

We implement our algorithm in \textbf{PyTorch}. \textbf{ResNet-50} \cite{cite:CVPR16DRL} is adopted as the feature extractor with parameters fine-tuned from the model pre-trained on ImageNet \cite{cite:Arxiv14ImageNet}. The main classifier and auxiliary classifier are both 2-layer neural networks with width 1024. For optimization, we use the mini-batch SGD with the Nesterov momentum 0.9. The learning rate of the classifiers are set 10 times to that of the feature extractor, the value of which is adjusted according to \citet{cite:JMLR16RevGrad}.

\begin{table}[tbp]
	\vspace{-5pt}
	\centering
	\addtolength{\tabcolsep}{-3pt}
	\caption{Accuracy (\%) on {VisDA-2017} (ResNet-50).}
	\label{table:visda}
	\vskip 0.05in
	\begin{tabular}{lc}
		\toprule
		Method                      & Synthetic $\rightarrow$ Real \\
		\midrule
		JAN \cite{cite:ICML17JAN}   & 61.6                         \\
		MCD \cite{cite:CVPR18MCD}   & 69.2                         \\
		GTA \cite{cite:CVPR18GTA}   & 69.5                         \\
		CDAN \cite{cite:NIPS18CDAN} & 70.0                         \\
		\textbf{MDD} (Proposed)               & \textbf{74.6}                \\
		\bottomrule
	\end{tabular}%
	\vspace{-10pt}
\end{table}

\subsection{Results}

The results on Office-31 are reported in Table~\ref{table:office31}. MDD achieves state of the art accuracies on five out of six transfer tasks. Notice that in previous works, feature alignment methods (JAN, CDAN) generally perform better for large-to-small tasks (A$\to$W, A$\to$D) while pixel-level adaptation methods (GTA) tend to obtain higher accuracy for small-to-large ones (W$\to$A, D$\to$A). Nevertheless our algorithm outperforms both types of methods on almost all task, showing its efficacy and universality.
Tables~\ref{table:officehome} and \ref{table:visda} present the accuracies of our algorithm on Office-Home and VisDA-2017, where we make remarkable performance boost. Some of the methods listed in the tables use additional techniques such as the entropy minimization to enhance their performance. Our method possesses both simplicity and performance strength.

\begin{figure*}[htbp]
	\vspace{-10pt}
	\centering
	\subfigure[Test Accuracy]{
		\includegraphics[width=0.23\textwidth]{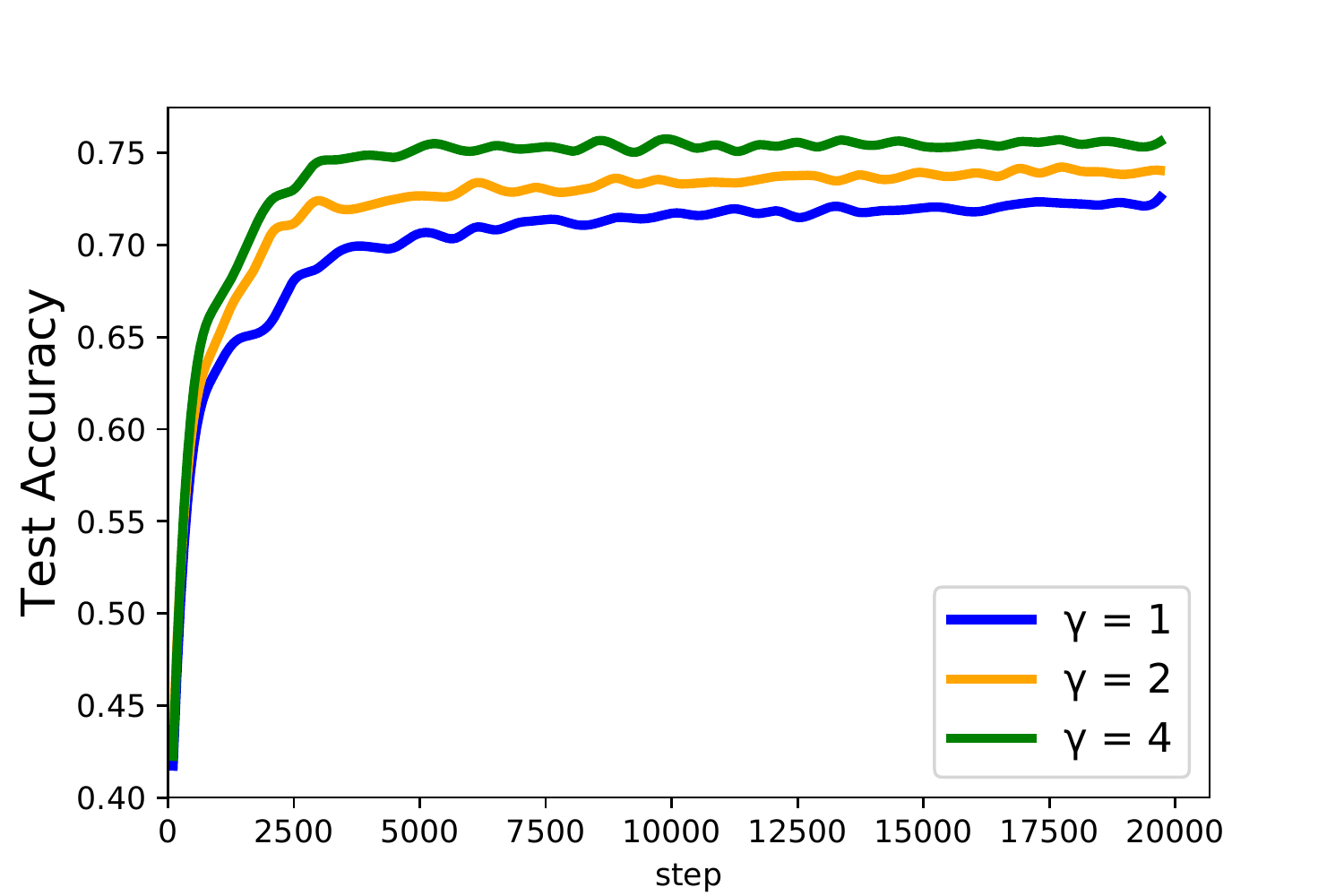}
		\label{fig:accuracy}
	}
	\hfil
	\subfigure[Equilibrium on Source]{
		\includegraphics[width=0.23\textwidth]{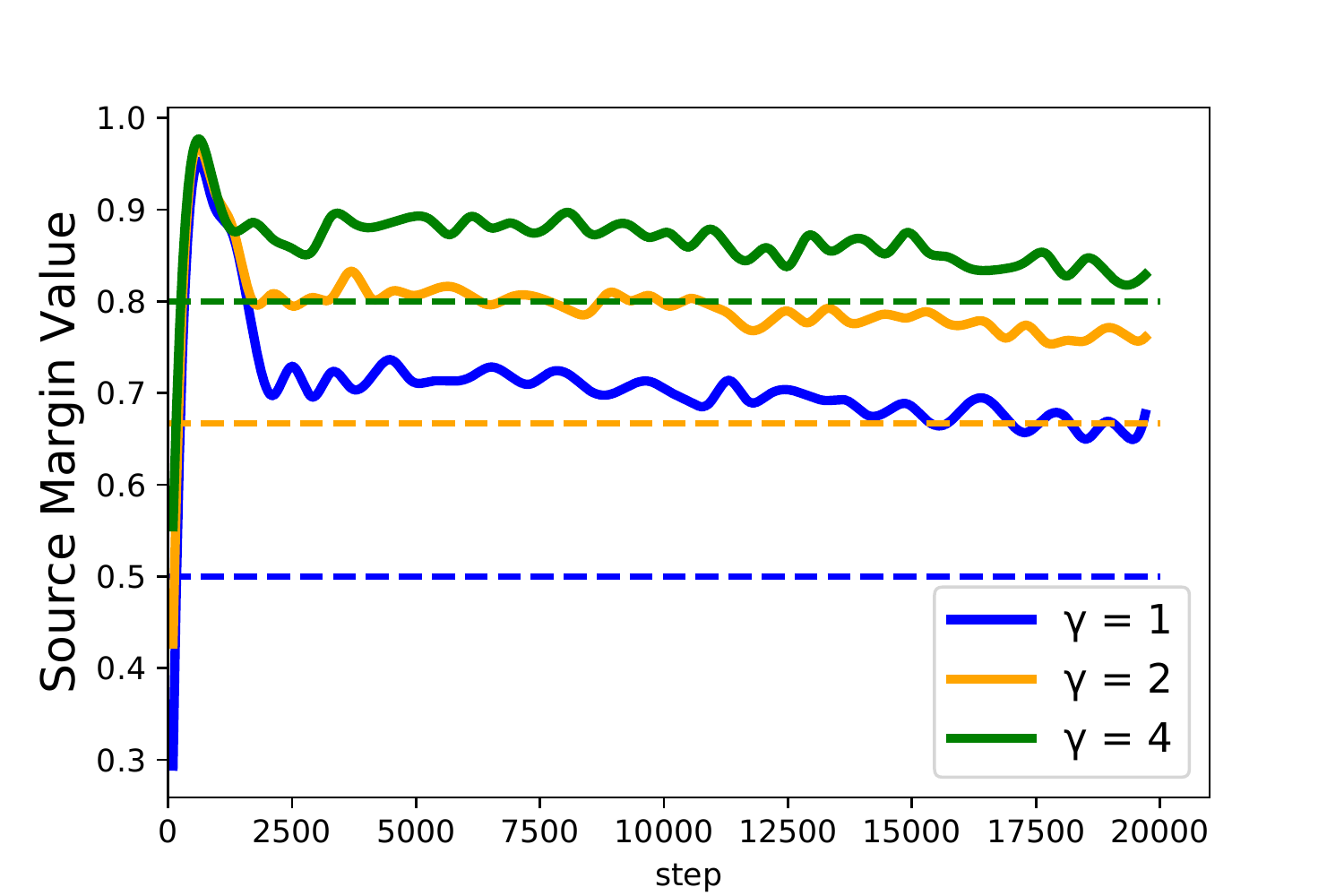}
		\label{fig:sourceeqvalue}
	}
	\hfil
	\subfigure[Equilibrium on Target]{
		\includegraphics[width=0.23\textwidth]{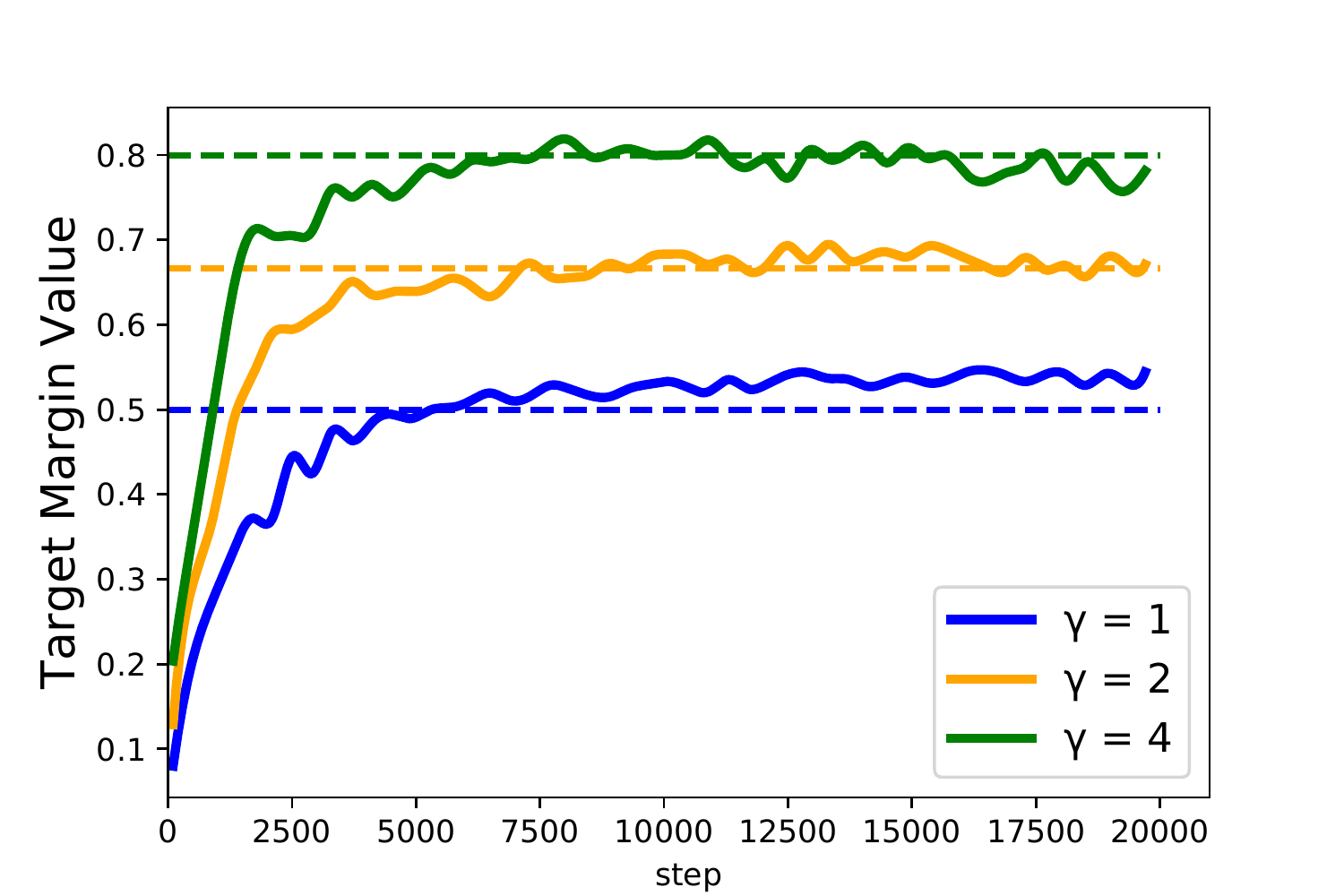}
		\label{fig:targeteqvalue}
	}
	\vspace{-10pt}
	\caption{Test accuracy and empirical values of $\sigma_{h_f}\comp f'$ on transfer task D $\rightarrow$ A, where dashed lines indicate \(\gamma/(1+\gamma)\).}
	\label{fig:acc}
	\vspace{-5pt}
\end{figure*}

\begin{figure*}[htbp]
	\vspace{-10pt}
	\centering
	\subfigure[MDD w/o Minimization]{
		\includegraphics[width=0.2\textwidth]{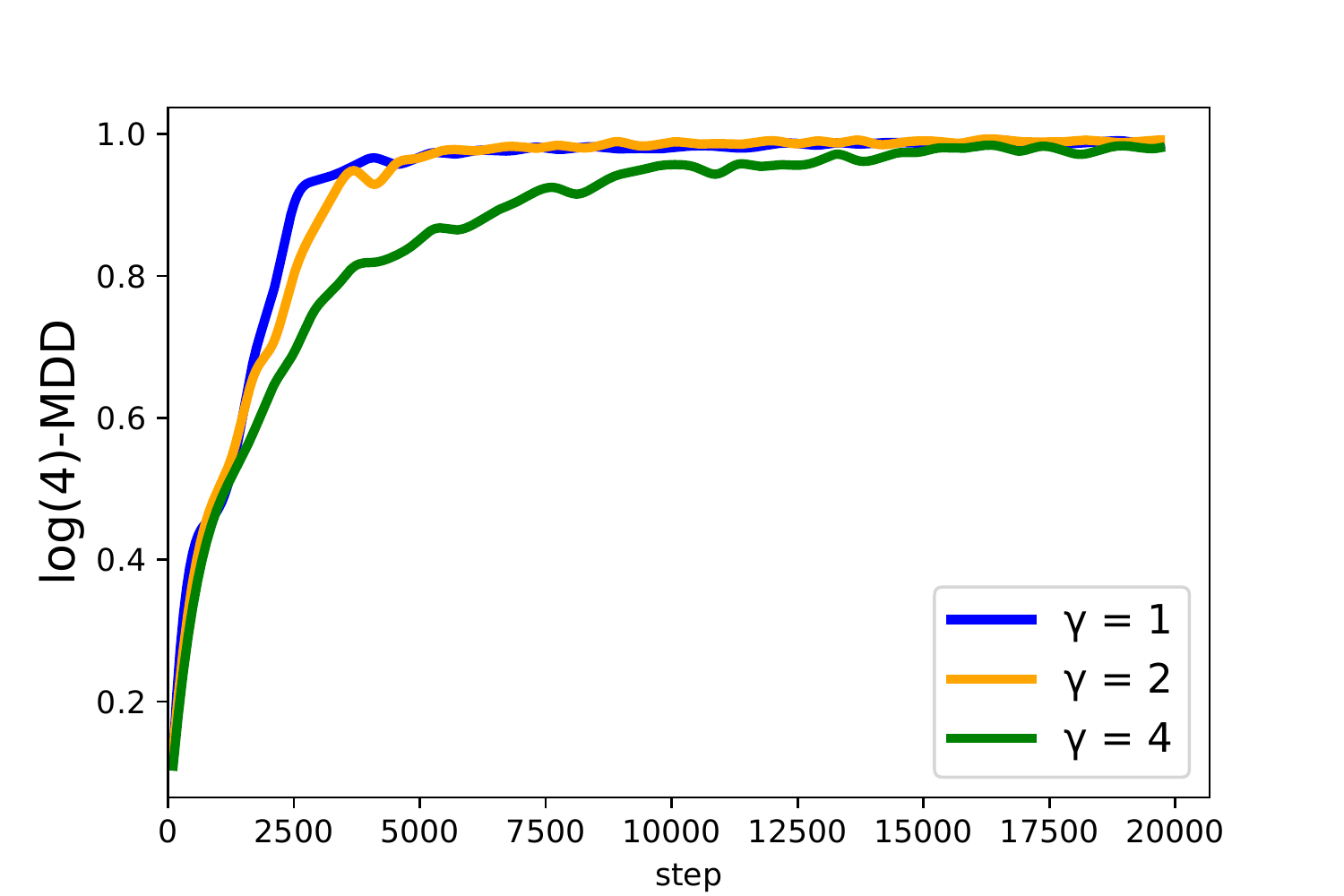}
		\label{fig:mdd0}
	}
	\hfil
	\subfigure[DD]{
		\includegraphics[width=0.2\textwidth]{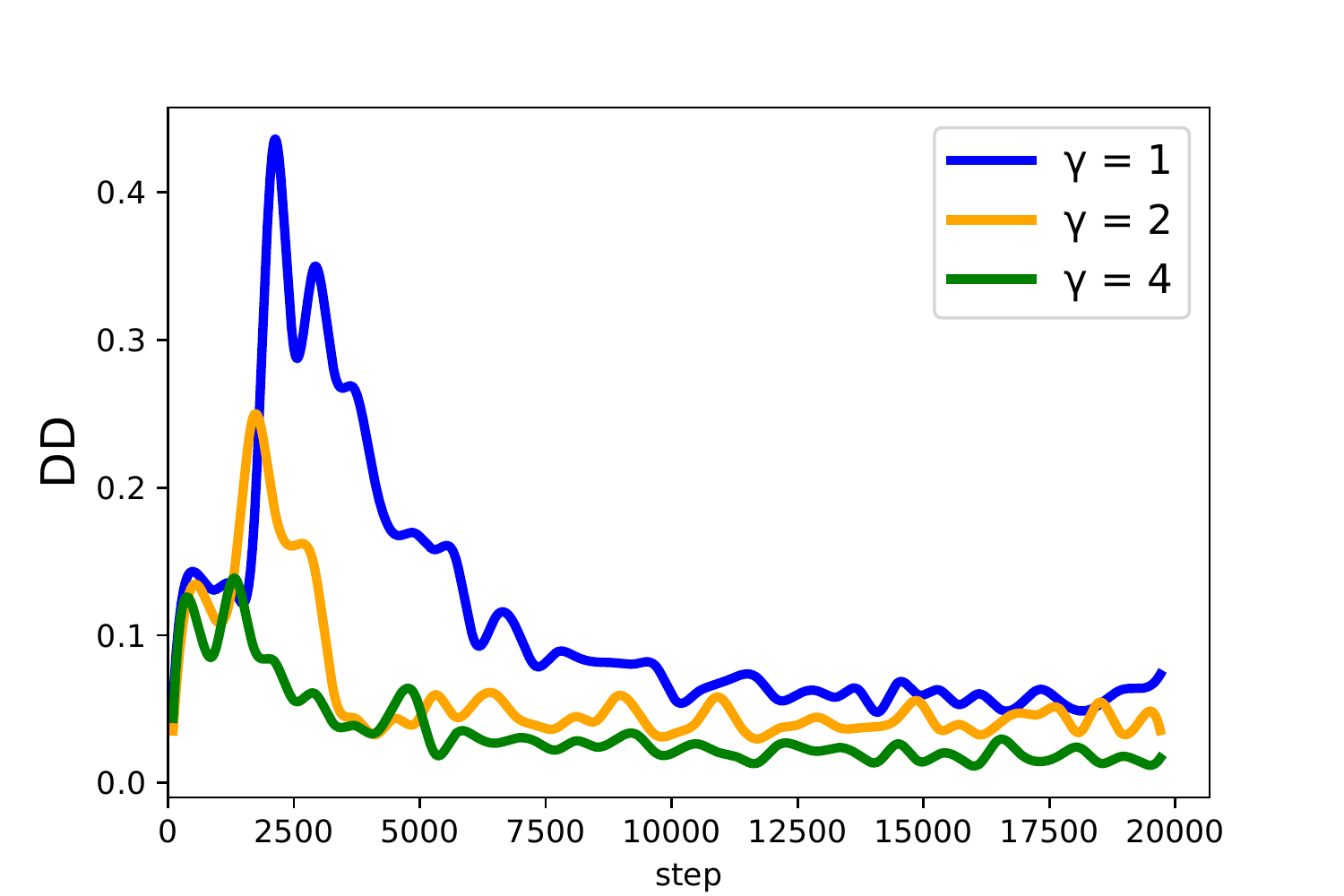}
		\label{fig:dd}
	}
	\hfil
	\subfigure[$\log 2$-MDD]{
		\includegraphics[width=0.2\textwidth]{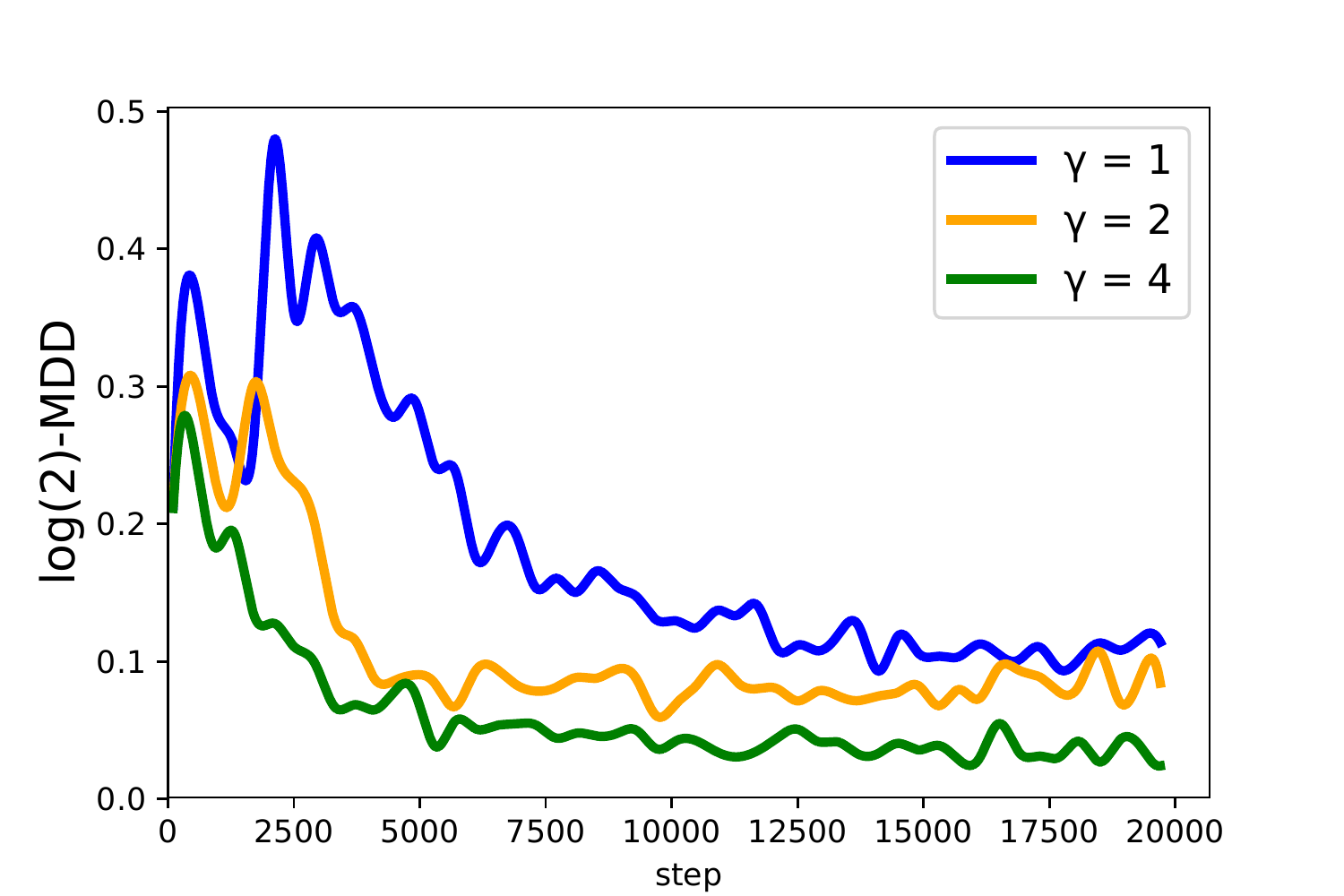}
		\label{fig:mdd2}
	}
	\hfil
	\subfigure[$\log 4$-MDD]{
		\includegraphics[width=0.2\textwidth]{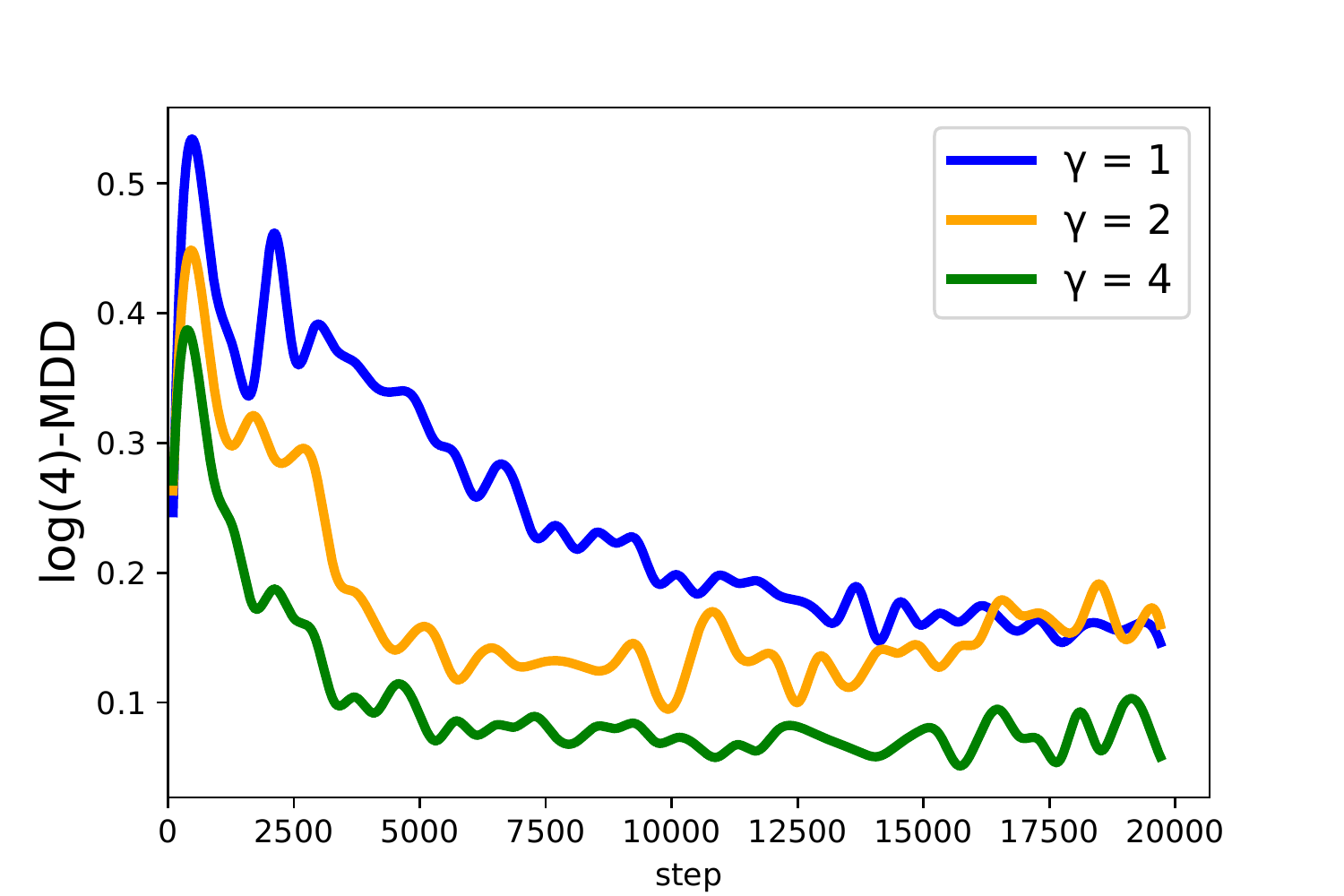}
		\label{fig:mdd4}
	}
	\vspace{-10pt}
	\caption{Empirical values of the margin disparity discrepancy (MDD) computed by auxiliary classifier $f'$.}
	\label{fig:mdd}
	\vspace{-5pt}
\end{figure*}

\subsection{Analyses}

In our adversarial learning algorithm, we reasonably use the combined cross-entropy loss instead of the margin loss and margin disparity discrepancy in our theory. We need to show that despite the technical modification, our algorithm can well reduce empirical MDD computed according to $f'$:
\begin{equation}
	\disp_{\widehat Q}^{(\rho)}(f',f)-\disp_{\widehat P}^{(\rho)}(f',f).
\end{equation}
We choose $\gamma  = 1, 2, 4$ for comparison. The expected margin should reach $\log 2$ and $\log 4$ in the last two cases while there is no guarantee for margin with $\gamma =1 $. Correspondingly, we examine DD (based on 0-1 loss), $\log 2$-MDD and $\log 4$-MDD for task D\(\to\)A and show results in Figures~\ref{fig:acc}--\ref{fig:mdd}.

\begin{table}[tbp]
	\vspace{-5pt}
	\centering
	\addtolength{\tabcolsep}{0pt}
	\caption{Accuracy (\%) on {Office-31} by different margins.}
	\label{table:gamma}
	\vskip 0.05in
	\begin{tabular}{cccc}
		\toprule
		Margin $\gamma$   & A $\rightarrow$ W & D $\rightarrow$ A & Avg on Office-31 \\
		\midrule
		1          & 92.5              & 72.4             & 87.6               \\
		2          & 93.7              & 73.0              & 88.1               \\
		3          & 94.0              & 73.7              & 88.5               \\
		\textbf{4} & \textbf{94.5}     & \textbf{74.6}     & \textbf{88.9}      \\
		5          & 93.8              & 74.3              & 88.7               \\
		6          & 93.5              & 74.2              & 88.6               \\
		\bottomrule
	\end{tabular}
	\vspace{-10pt}
\end{table}

First, we justify that without the minimization part of the adversarial training, the auxiliary classifier $f'$ in Eq.~\eqref{loss:gamma} is close to the $f'$ that maximizes MDD over $\mathcal{F}$. We solve this optimization problem by directly training with the auxiliary classifier and show our results in \ref{fig:mdd0}, where MDD reaches $1$ shortly after training begins, implying that the loss function we use can well substitute MDD.

Next, we consider the equilibrium of the minimax optimization. The average values of $\sigma_{h_f}\comp f'$ are presented in Figures~\ref{fig:sourceeqvalue} and \ref{fig:targeteqvalue}. We could see that at the final training stage, $\sigma_{h_f}\comp f'$ is close to the predicted value \(\gamma/(1+\gamma)\) on the target (Section \ref{exp:gamma}), which gives rise to large margin.

Last, by visualizing the values of DD, $\log 2$-MDD and $\log 4$-MDD and test accuracy computed over the whole dataset every 100 steps, we could see that relatively larger $\gamma$ leads to smaller MDD and higher test accuracy. Despite difficulties in gradient saturation, results using the original MDD loss are also comparable as shown in Appendix (See Table E.1).

\section{Related Work}
\paragraph{Domain Adaptation Theory.}

One of the pioneering theoretical works in this field was conducted by \citet{cite:NIPS07DAT}. They proposed the $\mathcal{H}\Delta \mathcal{H}$-divergence as a substitution of traditional distribution discrepancies (e.g. total variation, KL-divergence), which overcame the difficulties in estimation from finite samples. \citet{thy:mohri2009dd} considered a general class of loss functions satisfying symmetry and subadditivity and developed a generalization theory with respect to the newly proposed discrepancy distance.
The concurrent work in this setting was made by \citet{thy:kuroki2018sdisc}, who introduced a tractable and finer counterpart for \(\mathcal{H}\Delta \mathcal{H}\)-divergence called S-disc computed with the ideal source classifier and similar class of loss functions with \cite{thy:mohri2009dd}. In fact this measurement is encompassed in our DD as a special case.
\citet{thy:mohri2012ydsic, thy:ye2012ydsic} proposed $\mathcal{Y}$-disc for domain adaptation with partially labeled target data.
\citet{thy:cortes2014reg, thy:cortes2015gd} further proposed a theory for regression tasks in the setting of domain adaptation via the generalized discrepancy.
Another line of theoretical works on domain adaptation puts emphasis on the assumptions of the different distributions. \citet{cite:zhang2013domain,cite:gong2016domain} tackled this problem from a causal view and put forward the generalized target shift (GeTarS) scenario instead of the traditional assumption of covariate shift.
\citet{thy:germain2013bayes} proposed a PAC-Bayesian theory for domain adaptation using the domain disagreement pseudometric.

\vspace{-10pt}

\paragraph{Domain Adaptation Algorithm.}

Domain adaptation methods based on deep networks have achieved great success in recent years \cite{cite:ICML15DAN, cite:ICML15RevGrad}. These works aim to learn domain-invariant representations by minimizing a certain discrepancy between distributions of source and target features extracted by a shared representation learner.
With insights from both the theory of \citet{thy:shai10ad} and the practice of adversarial learning \cite{cite:NIPS14GAN}, \citet{cite:ICML15RevGrad} put forward the domain adversarial neural network (DANN). A domain discriminator is trained to distinguish source features from target features and a feature extractor to confuse the discriminator. 
Since then, a series of works have appeared and achieved significantly better performance.
\citet{cite:CVPR17ADDA} proposed an architecture that employed asymmetric encodings for target and source data.
\citet{cite:NIPS18CDAN} presented a principled framework that conducted the adversarial adaptation models using conditional information.
\citet{cite:ICML18CyCADA, cite:CVPR18GTA} unified pixel-level and feature-level adversarial learning for domain adaptation.
\citet{cite:CVPR18MCD} considered the classifiers instead of features and designed an original adversarial learning method by maximizing the classifier discrepancy.

\vspace{-5pt}

\section{Conclusion}
In this paper, we derived novel generalization bounds based on newly proposed margin disparity discrepancy, and presented both theoretical and algorithmic analyses of domain adaptation.
Our analyses are more general for analyzing real-world domain adaptation problems, and the well-designed theory-induced algorithm achieves the state of the art results.

\section*{Acknowledgements}
This work was supported by the National Natural Science Foundation of China (61772299, 71690231, and 61672313).

\bibliography{paper}
\bibliographystyle{icml2019}
\end{document}


\twocolumn[

	\icmltitle{Supplemental Material:\\Bridging Theory and Algorithm for Domain Adaptation}

	\icmlsetsymbol{equal}{*}

	\begin{icmlauthorlist}
		\icmlauthor{Yuchen Zhang}{equal,ss,lab}
		\icmlauthor{Tianle Liu}{equal,lab,math}
		\icmlauthor{Mingsheng Long}{ss,lab}
		\icmlauthor{Michael I. Jordan}{ucb}
	\end{icmlauthorlist}

	\icmlaffiliation{ss}{School of Software}
	\icmlaffiliation{math}{Department of Mathematical Science, Tsinghua University, China}
	\icmlaffiliation{lab}{Research Center for Big Data, BNRist}
	\icmlaffiliation{ucb}{University of California, Berkeley, USA.
	
	$^\dag$Yuchen Zhang <zhangyuc17@mails.tsinghua.edu.cn>}
	\icmlcorrespondingauthor{Mingsheng Long}{mingsheng@tsinghua.edu.cn}
	\icmlkeywords{Machine Learning, ICML}

	\vskip 0.3in
]

\printAffiliationsAndNotice{\icmlEqualContribution}

\section{Properties of DD}

\begin{proposition}
Let \(\mathcal P\) denote the space of probability distributions over the domain \(\mathcal X\). For any \(h\in \mathcal H\) in the setting of binary classification, the induced Disparity Discrepancy \(d_{h,\mathcal H}(\cdot,\cdot)\) is a pseudometric on \(\mathcal P\). More precisely it is a metric on some quotient space of \(\mathcal P\).
\end{proposition}

\begin{proof}
Firstly, \(d_{h,\mathcal H}(P,Q)\) is non-negative since \(h\in\mathcal H\) and \(d_{h,\mathcal H}(P,P)=0\) holds for any \(P\in\mathcal P\) by definition.

Secondly, \(d_{h,\mathcal H}(P,Q)\) is symmetric. Otherwise suppose \(d_{h,\mathcal H}(P,Q)>d_{h,\mathcal H}(Q,P)\). Therefore, we can choose \(g\in \mathcal H\) such that \(\mathbb{E}_Q \mathbb{1}[g\ne h] - \mathbb{E}_P \mathbb{1}[g\ne h]>d_{h,\mathcal H}(Q,P)\). By our assumption, \(1-g\in\mathcal H\). Thus
\[\begin{aligned}
&d_{h,\mathcal H}(Q,P)
\\ 
\geq&\ex_P\mathbb 1[1-g\neq h]-\ex_Q\mathbb 1[ 1-g\neq h]
\\
=&\ex_Q\mathbb 1[g\neq h]-\ex_P\mathbb 1[g\neq h]
\\
>& d_{h,\mathcal H}(Q,P).\end{aligned}\]
Contradiction.

Lastly, for any distribution \(P,Q,R\) we have
\[\begin{aligned}&d_{h,\mathcal H}(P,Q)\\
 = &\sup_{h'\in \mathcal{H}}(\disp_Q(h',h) - \disp_P(h',h) )\\ 
\leq &\sup_{h'\in \mathcal{H}}(\disp_Q(h',h) - \disp_R(h',h) )\\ 
& \qquad+\sup_{h''\in \mathcal{H}}(\disp_R(h'',h) - \disp_{P}(h'',h) ) \\ 
=& d_{h,\mathcal H}(R,Q)+d_{h,\mathcal H}(R,P).\end{aligned}\]
Thus \(d_{h,\mathcal H}(P,Q)\) is a pseudometric on \(\mathcal P\).

Note that \(d_{h,\mathcal H}(P,Q)= 0\) does not imply \(P=Q\) in general. However, this equation gives a equivalence relation \(\sim_{h,\mathcal H}\) on \(\mathcal P\), which could be easily checked noticing that \(d_{h,\mathcal H}(P,Q)=0\) is equivalent to \[\ex_Q\mathbb 1[h'\neq h]=\ex_P\mathbb 1[h'\neq h]\] for any \(h'\in \mathcal H\). Thus \(d_{h,\mathcal H}(\cdot,\cdot)\) is a metric on the quotient space \(\mathcal P/{\sim_{h,\mathcal H}}\).
\end{proof}




Next in the binary classification setting we show that there are strong connections between the \(\mathcal{H}\Delta\mathcal{H}\)-distance and disparity discrepancy. For a hypothesis set \(\mathcal H\), the \emph{symmetric difference hypothesis set} \(\mathcal H\Delta \mathcal H\) is the set of classifiers
\begin{equation}\mathcal{H}\Delta\mathcal{H}\triangleq\{h-h'|h,h'\in\mathcal{H} \}.\end{equation}
\begin{proposition}
For the binary classification, \begin{equation}d_{\mathcal H \Delta \mathcal H}(P,Q)=\sup_{h\in\mathcal H} d_{h,\mathcal H}(P,Q).\end{equation}
\end{proposition}

\begin{proof}
By definition,
\[\begin{aligned}\sup_{h\in\mathcal H} d_{h,\mathcal H}(P,Q)& =\sup_{h,h'\in\mathcal H}(\disp_Q(h',h) - \disp_P(h',h) )\\
& =\sup_{g\in\mathcal H\Delta\mathcal H}(\mathbb{E}_Q \mathbb{1}[g\ne 0] - \mathbb{E}_P \mathbb{1}[g\ne 0] )\\ & =\sup_{g\in\mathcal H\Delta\mathcal H}(\mathbb{E}_Q g- \mathbb{E}_P g)\\ & =d_{\mathcal H \Delta \mathcal H}(P,Q).\end{aligned}\]
\end{proof}
Now we consider when our proposed discrepancy is independent of the selection of \(h\), in which case the disparity discrepancy is actually equivalent to \(\mathcal H\Delta \mathcal H \)-distance. A sufficient condition is stated below:

\begin{proposition}
For the binary classification, if the hypothesis set \(\mathcal H\) is a linear space over the prime field \(\mathbb Z_2\), in other words \(\mathcal H\Delta \mathcal H=\mathcal H\), we have
\begin{equation}d_{h,\mathcal H}(P,Q)=d_{\mathcal H\Delta\mathcal H}(P,Q)=d_{\mathcal H}(P,Q)\end{equation}
	for any \(h\in\mathcal H\).
\end{proposition}

\begin{proof} Suppose there exist \(h,h'\in \mathcal H\) such that \[d_{h,\mathcal{H}}(P,Q)>d_{h',\mathcal H}(P,Q),\]
Then by the definition of \(d_{h,\mathcal H}(P,Q)\), for any \(\epsilon>0\) there exists \(g\in\mathcal H\) such that \[d_{h,\mathcal H}(P,Q)-(\ex_Q\mathbb 1[g\neq h]-\ex_P\mathbb 1[g\neq h])<\epsilon.\]
Let \(\epsilon=\frac 1 2 (d_{h,\mathcal H}(P,Q)-d_{h',\mathcal H}(P,Q))\), then consider \(g'=g-h+h'\)
\[
\begin{aligned}&\ex_Q\mathbb 1[g'\neq h']-\ex_P\mathbb 1[g'\neq h']\\=& \ex_Q\mathbb 1[g\neq h]-\ex_P\mathbb 1[g\neq h]\\>&d_{h',\mathcal H}(P,Q).\end{aligned}
\]
Contradiction.
\end{proof}

However, in the case of neural networks, especially with activations such as the rectified linear unit (ReLU), the condition mentioned above is generally not satisfied and \(\mathcal H\Delta \mathcal H \)-distance is often strictly larger than ours. To verify this, we provide the following example:

\begin{example}
Let \(\mathcal X=\mathbb R^2\) and \(P, Q\) be two dirac masses on the points \((-1,1)\) and \((1,-1)\) respectively. Let \(a,b\) be two parameters with values in \(\mathbb R\). Let \(r(t)\triangleq\max\{0,t\}\) be the ReLU function. For any input data \(x=(x_1,x_2)\in \mathbb R^2\), the pseudo label predicted by a hypothesis \(h\in\mathcal H\) is defined as follows:
\[h(x)=\begin{cases}1& \mathrm{if}\ r(x_1-a)\geq r(x_2-b)\\0 & \mathrm{if}\ r(x_1-a)<r(x_2-b)\end{cases}.\]

One can check that there are three kinds of hypotheses with values \[h_1(P,Q)=(0,0), h_2(P,Q)=(0,1),h_3(P,Q)=(1,1).\] Then \(d_{h_1,\mathcal H}(P,Q)\!=\!1\) and \(d_{h_3,\mathcal H}(P,Q)\!=\!0\), in which case \(d_{h_3,\mathcal H}(P,Q)\) does not coincide with \(d_{\mathcal H\Delta \mathcal H}(P,Q)\).
\end{example}
\section{Generalization Bounds with DD}

\begin{lemma}[{Rademacher Generalization Bound, Theorem 3.1 of \citet{thy:mohri2012foundations}}]\label{2.1}
\it Suppose that $\mathcal G$ is a class of function maps $\mathcal{X} \to [0,1]$. For any $\delta > 0$, with probability at least $1-\delta$, the following holds for all $g\in\mathcal G$:
	\begin{equation}
	| \ex_{D} \, g - \ex_{\widehat D}\, g| \leq 2\mathfrak R_{n,D}(\mathcal{G}) + \sqrt{\frac{\log\frac{2}{\delta}}{2n}}.
	\end{equation}
\end{lemma}

\begin{theorem}\label{2.2}
	For any classifier $h$  
		\begin{equation} 
		\err_{Q} (h) \le \err_P (h) +  d_{h,\mathcal{H}}(P,Q) +    \lambda  ,
		\end{equation} 
		where 
		\(\lambda=\lambda(\mathcal{H},P,Q) \) is independent of \(h\).
\end{theorem}
	
\begin{proof} Let \(h^*\) be the ideal joint classifier which minimizes the combined error, 
	\[ h^*\triangleq\mathop{\arg\min}\limits_{h\in \mathcal{H}} \{\err_P(h)+\err_Q(h)\}.\]
	Set \(\lambda =\err_P (h^*)+\err_Q(h^*)\). Then
	\begin{align*} 
	&\err_{Q} (h) = \err_{P} (h) + \err_{Q} (h) -\err_{P} (h)
	\\
	\le &\err_P(h)
	+ 
	(\mathbb{E}_Q \mathbb{1}[h^*\ne h] - \mathbb{E}_P \mathbb{1}[h^*\ne h] ) 
	\\
	&+ (\err_{P} (h^*) + \err_{Q} (h^*))
	\\
	\le &\err_P(h)
	+    \sup_{h'\in \mathcal{H}}   
	(\disp_Q(h',h) - \disp_P(h',h) )
	+    \lambda  \\
	=& \err_P (h) +  d_{h,\mathcal{H}}(P,Q) +    \lambda .
	\end{align*}
\end{proof}

\begin{theorem}
Suppose $\mathcal{H}$ is a hypothesis space maps $\mathcal{X}$ to $\{0,1\}$. $\widehat D$ is empirical distribution corresponding to datasets contains $n$ data points sampled from $D$. For any $\delta > 0$, with probability at least $1-\delta$, the following holds for all \(h, h'\in\mathcal H\):
		\begin{equation} \begin{aligned}
		|\disp_D(h',h)-&\disp_{\widehat D}(h',h)| 
		\\
		&
		\leq 2\mathfrak R_{n, D}(\mathcal{H} \Delta \mathcal{H} ) + \sqrt{\frac{\log\frac{2}{\delta}}{2n}}.
		\end{aligned}
		\end{equation} 
\end{theorem}
\begin{proof}
	\[\begin{aligned}
	&\sup_{h, h'\in \mathcal{H}}   
	| \mathbb{E}_{\widehat{D}} \mathbb{1}[h'\ne h]  -\mathbb{E}_D \mathbb{1}[h'\ne h] |
	\\
	=&\sup_{g\in \mathcal{H}\Delta\mathcal{H}}   
	| \mathbb{E}_{\widehat{D}} \mathbb{1}[g\ne \mathbf 1]  -\mathbb{E}_D \mathbb{1}[g\ne \mathbf 1] |
	\\
	=&\sup_{g\in \mathcal{H}\Delta\mathcal{H}}   
	| \mathbb{E}_{\widehat{D}} g -\mathbb{E}_D g|.
	\end{aligned}\]
	With Lemma \ref{2.1}, we could know that 
		\[
	\sup_{g\in \mathcal{H}\Delta\mathcal{H}} | \ex_{D} \, g - \ex_{\widehat D}\, g| \leq 2\mathfrak R_{n, D}(\mathcal{H}\Delta\mathcal{H}) + \sqrt{\frac{\log\frac{2}{\delta}}{2n}}.
	\]
\end{proof}

\begin{theorem}
For any $\delta>0$ and binary classifier $h\in \mathcal{H}$, with probability $1-3\delta$, we have
	\begin{equation}
	\begin{aligned}
	\err_Q(h)&\le \err_{\widehat{P}}(h) + d_{h,\mathcal{H}}(\widehat{P},\widehat{Q}) +\lambda
	\\
	&+2\mathfrak R_{n, P}(\mathcal{H} \Delta \mathcal{H} )+2\mathfrak R_{n, P}(\mathcal{H})  + 2\sqrt{\frac{\log\frac{2}{\delta}}{2n}} 
	\\
	&+ 2\mathfrak R_{m, Q}(\mathcal{H} \Delta \mathcal{H} ) + \sqrt{\frac{\log\frac{2}{\delta}}{2m}}.\end{aligned}\end{equation}
where $\lambda  = \min_{h'\in \mathcal{H}} (\err_P(h') + \err_Q(h'))$ is independent with $h$.
\end{theorem}

\begin{proof} Consider the difference of expected and empirical terms on the right-hand side.
\[\begin{aligned}
&\sup_{h\in \mathcal{H}}(\err_P(h) + d_{h,\mathcal{H}}(P,Q)-\err_{\widehat{P}}(h) - d_{h,\mathcal{H}}(\widehat{P},\widehat{Q}))
\\
=&\sup_{h\in \mathcal{H}}(\err_P(h)-\err_{\widehat{P}}(h) + d_{h,\mathcal{H}}(P,Q) - d_{h,\mathcal{H}}(\widehat{P},\widehat{Q}))
\\
\le&\sup_{h\in \mathcal{H}}(\err_{\!P}\!(h)\!-\!\err_{\!\widehat{P}}\!(h)\!) + \sup_{h\in \mathcal{H}}(d_{h,\!\mathcal{H}}\!(P,\!Q)\! -\! d_{h,\!\mathcal{H}}\!(\widehat{P},\!\widehat{Q})\!).
\end{aligned}\]

First by Lemma \ref{2.1},  $\forall\  \delta > 0$, with probability $1-\delta$,
\[
\sup_{h\in \mathcal{H}}(\err_P(h)-\err_{\widehat{P}}(h)) \le 2\mathfrak R_{n, P}(\mathcal{H})  + \sqrt{\frac{\log\frac{2}{\delta}}{2n}} .
\]

Then we bound the difference between $d_{h,\mathcal{H}}(P,Q) $ and \(d_{h,\mathcal{H}}(\widehat{P},\widehat{Q})\):
\[\begin{aligned}
&\quad d_{h,\mathcal{H}}(P,Q) - d_{h,\mathcal{H}}(\widehat{P},\widehat{Q})
\\ \\
=&  \sup_{h'\in \mathcal{H}}   
(\disp_Q(h',h) - \disp_P(h',h) ) 
\\
&\qquad -\sup_{h''\in \mathcal{H}}   
(\disp_{\widehat Q}(h'',h) -\disp_{\widehat P}(h'',h) )
\\
\le & \sup_{h'\in \mathcal{H}}   
\big(\disp_Q(h',h) - \disp_P(h',h) 
\\
&\qquad -\disp_{\widehat Q}(h',h)+\disp_{\widehat P}(h',h) \big)
\\
\le&  \sup_{h'\in \mathcal{H}}   
(\disp_Q(h',h) - \disp_{\widehat Q}(h',h)) 
\\
&\qquad +\sup_{h''\in \mathcal{H}}   
( \disp_{\widehat P}(h'',h)  -\disp_{P}(h'',h) ).
\end{aligned}\]

Take supremum over $ h \in \mathcal{H}$, we have:
\[\begin{aligned}
&\sup_{h\in \mathcal{H}} (d_{h,\mathcal{H}}(P,Q) - d_{h,\mathcal{H}}(\widehat{P},\widehat{Q}))
\\
\le&  \sup_{h, h'\in \mathcal{H}}   
|\disp_Q(h',h) - \disp_{\widehat Q}(h',h)| +
\\
&\sup_{h, h''\in \mathcal{H}}   
| \disp_{\widehat P}(h'',h)  -\disp_{P}(h'',h) |.
\end{aligned}\]

From Theorem \ref{2.2}, we directly get:
\[\begin{aligned}
&\sup_{h\in \mathcal{H}} (d_{h,\mathcal{H}}(P,Q) - d_{h,\mathcal{H}}(\widehat{P},\widehat{Q}))
\\
\le&  2\mathfrak R_{n, P}(\mathcal{H} \Delta \mathcal{H} )+2\sqrt{\frac{\log\frac{2}{\delta}}{2n}} + 2\mathfrak R_{m, Q}(\mathcal{H} \Delta \mathcal{H} ) + \sqrt{\frac{\log\frac{2}{\delta}}{2m}}.
\end{aligned}\]
Combine the two parts of inequality, we get the final result. \end{proof}

We introduce theory of VC-dimension here to further measure the generalization ability.

\begin{definition}[VC-Dimension]
 The VC-dimension of a hypothesis set \(\mathcal H\) is the size of the largest set that can be fully shattered by \(\mathcal H\). 
	Let \begin{equation}\Pi(n,\mathcal H)\triangleq\max_{x_1,\ldots,x_n}\big{|}\{h(x_1),\ldots,h(x_n)\big {|}h\in\mathcal H\}\big{|}.\end{equation} Then
	\begin{equation}\mathrm{VC}(\mathcal H)\triangleq\max \{m\big{|}\Pi(m,\mathcal H)=2^m\}.\end{equation}
\end{definition}

\begin{lemma}[Corollary 3.1 \& 3.3 of \citet{thy:mohri2012foundations}]\label{2.6}
 Suppose $\mathcal{G}$ takes value in $\{0,1 \}$ and \(d\) is the VC-dimension of \(\mathcal G\). Then the Rademacher complexity of $\mathcal{G}$ has the following holds for all \(h\in\mathcal H\),
	\begin{equation}
	\mathfrak R_{n, D}(\mathcal{G} ) \le \frac{1}{2}\sqrt{\frac{2d\log \frac{en}{d}}{n}}.
	\end{equation}
\end{lemma}

\begin {theorem}
	For any $\delta>0$ and $h\in \mathcal{H}$, with probability $1-3\delta$, we have\begin{equation}
	\begin{aligned}
	\err_Q(h)&\le \err_{\widehat{P}}(h) + d_{h,\mathcal{H}}(\widehat{P},\widehat{Q}) +\lambda
	\\&\quad+C_1\sqrt{ \frac{d\log\frac{\mathrm en}{d}}{n}} + C_2\sqrt{\frac{\log\frac{2}{\delta}}{2n}} 
	\\&
	+ C_3\sqrt{ \frac{d\log\frac{\mathrm em}{d}}{m}} + C_4\sqrt{\frac{\log\frac{2}{\delta}}{2m}}+\lambda,
	\end{aligned}\end{equation}
	where \(C_1,C_2\) are constants independent of \(\mathcal H, P,Q\).
\end{theorem}
\begin{proof}
Since we can represent every \(g\in\mathcal H\Delta\mathcal H\) as a linear threshold network of depth \(2\) with \(2\) hidden units, the VC-dimension of \(\mathcal H \Delta \mathcal H\) is at most twice the VC-dimension of \(\mathcal H\) \cite{thy:anthony2009neural}. Let \(g\triangleq 1 +h-h'\), then \(g\in\mathcal H\Delta \mathcal H\) and \(h\neq h'\) is equivalent to \(g\neq 1\). Thus by Lemma \ref{2.6} we have

\[\mathfrak R_{n, P}(\mathcal{H}\Delta \mathcal{H} ) 
\le\ \frac 1 2\sqrt{ \frac{2d\log\frac{\mathrm en}{4d}}{n}}.\]

To summarize, with probability $1-3\delta$,
\[\begin{aligned}
\err_Q(h)&\le \err_P(h)+d_{h,\mathcal{H}}(P,Q) +\lambda
\\
&\le \err_{\widehat{P}}(h) + d_{h,\mathcal{H}}(\widehat{P},\widehat{Q}) \\&\quad+4\sqrt{ \frac{d\log\frac{\mathrm en}{d}}{n}} + 2\sqrt{\frac{\log\frac{2}{\delta}}{2n}}+\\&\quad+2\sqrt{ \frac{d\log\frac{\mathrm em}{d}}{m}} + \sqrt{\frac{\log\frac{2}{\delta}}{2m}}+\lambda. 
\end{aligned}\]
\end{proof}

\section{Generalization Bounds with MDD}
	
\begin{lemma}\label{the:triangle}
For any distribution $D$ and any $f$, we have 
	\begin{equation}
	\disp_D^{(\rho)}(f',f) \le  \err_{D}^{(\rho)}(f') + \err_{D}^{(\rho)}(f).
	\end{equation}
\end{lemma}
	
\begin{proof}
We prove that for any $(x_i, y_i)$,
\[
\Phi_\rho\comp\rho_{f'}(x_i,h_f(x_i)) \le \Phi_\rho\comp\rho_{f'}(x_i,y_i)  + \Phi_\rho\comp\rho_{f}(x_i,y_i)  ,
\]
If $h_f(x_i) \ne y_i$ or $h_{f'}(x_i) \ne y_i$, the right side of above equation will reach $1$, which is a trivial upper bound for the left part.
Otherwise $h_f(x_i) = h_{f'}(x_i) = y_i$, and 
\begin{align*}
&\Phi_\rho\comp\rho_{f'}(x_i,h_f(x_i)) 
\\
\le & \Phi_\rho\comp\rho_{f'}(x_i,h_f(x_i)) + \Phi_\rho\comp\rho_{f}(x_i,y_i) 
\\
= & \Phi_\rho\comp\rho_{f'}(x_i,y_i) + \Phi_\rho\comp\rho_{f}(x_i,y_i).
\end{align*}
Take expectation on distribution $P$ and we get the result.
\end{proof}
\begin{theorem}[Proposition 3.3]\label{the:bound1}
	For any scoring function \(f\),
		\begin{equation}
		\err_{Q} (h_f) \le \err_P^{(\rho)} (f) +  d_{f,\mathcal{F}}^{(\rho)}(P,Q) +    \lambda,
		\end{equation}
		where 
		\(\lambda=\lambda(\rho,\mathcal{F},P,Q) \) is a constant independent of \(f\).
\end{theorem}
\begin{proof}
	Let \(f^*\) be the ideal joint hypothesis which minimizes the combined margin loss, 
	\[f^*\triangleq\mathop{\arg\min}\limits_{f\in \mathcal{H}} \{\err_P^{(\rho)} (f)+\err_Q^{(\rho)}(f)\}.\]
	Set \(\lambda =\err_P^{(\rho)} (f^*)+\err_Q^{(\rho)}(f^*)\). Then by Lemma~\ref{the:triangle},
	\begin{align*} 
	\err_{Q} (f) \le & \; \mathbb{E}_Q \mathbb{1}[h_{f}\ne h_{f^*}] +  \mathbb{E}_Q \mathbb{1}[h_{f^*}\ne y]
	\\
	\le& \;\err_{P}^{(\rho)}(f) -\err_{P}^{(\rho)} (f) 
	\\
	&+ \disp_Q^{(\rho)}(f^*,f) + \err_{Q}^{(\rho)} (f^*)
	\\
	\le& \;\err_{P}^{(\rho)}(f) + \err_{P}^{(\rho)} (f^*) - \disp_P^{(\rho)}(f^*,f) 
	\\
	&+  \disp_Q^{(\rho)}(f^*,f)   + \err_{Q}^{(\rho)} (f^*)
	\\
	\le & \;\err_P^{(\rho)} (f) +  d_{f,\mathcal{F}}^{(\rho)}(P,Q) + \lambda
	\end{align*}
	\vskip -20pt
\end{proof}

\begin{definition}
	Given a class of scoring functions \(\mathcal F\) and a class of the induced classifiers \(\mathcal H\), we define $\Pi_\mathcal{H}\mathcal{F}$ as
	\begin{equation}
		\Pi_{\mathcal H}\mathcal{F} = \{x \mapsto f(x, h(x)) | h\in \mathcal{H}, f\in \mathcal{F} \}.
	\end{equation}
\end{definition}

There is a geometric interpretation of the set \(\Pi_{\mathcal H}\mathcal F\) \cite{galbis2012vector}. Assuming \(\mathcal X\) is a manifold, assigning a vector space \(\mathbb R^k\) to each point in \(\mathcal X\) yields a vector bundle \(\mathcal B\). Now regarding the values of \(\mathcal H\) as one-hot vectors in \(\mathbb R^k\), \(\mathcal F\) and \(\mathcal H\) are both sets of sections of \(\mathcal B\) containing (probably piecewise continuous) vector fields. \(\Pi_{\mathcal H}\mathcal F\) can be seen as the space of inner products of vector fields from \(\mathcal H\) and \(\mathcal F\),
\begin{equation}\Pi_{\mathcal H}\mathcal F=\langle \mathcal H,\mathcal F\rangle=\{\langle h,f\rangle\big{|}h\in \mathcal H,f\in\mathcal F\}.\end{equation}

\begin{lemma}[A modified version of Theorem 8.1, \citet{thy:mohri2012foundations}]\label{3.1}
Suppose $\mathcal F\subseteq \mathbb R^{\mathcal X\times \mathcal Y}$ is the hypothesis set of scoring functions with $\mathcal Y=\{1,2,\ldots ,k\}$. Let
	\begin{equation}
	\Pi_1\mathcal F\triangleq\{x\mapsto f(x,y)\big{|}y\in \mathcal Y,f\in \mathcal F\}.
	\end{equation}
	Fix $\rho>0$. Then for any $\delta>0$, with probability at least $1-\delta$, the following holds for all $f\in\mathcal F$:
	\begin{equation}
	|\\err_D^{(\rho)}(f)-\\err_{\widehat D}^{(\rho)}(f)|\leq\frac{2k^2}{\rho}\mathfrak{R}_{n, D}(\Pi_1\mathcal F)+\sqrt{\frac{\log \frac 2 {\delta}}{2n}}.
	\end{equation}
\end{lemma}

Note that a simple corollary of this lemma is the margin bound for multi-class classification:
\begin{equation}\label{equation18}
\begin{aligned}\err_D(h_f)&\leq \err_D^{(\rho)}(f)\\ & \leq \err_{\widehat D}^{(\rho)}(f)+\frac{2k^2}{\rho}\mathfrak{R}_{n,D}(\Pi_1\mathcal F)+\sqrt{\frac{\log \frac 2 {\delta}}{2n}}.\end{aligned}
\end{equation}

\begin{lemma}[Talagrand's lemma, \citet{thy:talagrand2014upper,thy:mohri2012foundations}]\label{3.2}
Let $\Phi: \mathbb{R} \to \mathbb{R}$ be an $\ell$-Lipschitz. Then for any hypothesis set $\mathcal{F}$ of real-valued functions, and any sample $\widehat D$ of size $n$, the following inequality holds:
\begin{equation}
\widehat{\mathfrak{R}}_{\widehat D}(\Phi \,\comp\,\mathcal F)  \le\ell \,\widehat{\mathfrak{R}}_{\widehat D}(\mathcal F) 
\end{equation}
\end{lemma}

\begin{lemma}[Lemma 8.1 of \citet{thy:mohri2012foundations}]\label{3.3}
Let $\mathcal{F}_1,\ldots,\mathcal{F}_k$ be $k$ hypothesis sets in $\mathbb{R}^{\mathcal{X}}$, $k>1$. 
	$\mathcal{G} = \{\max \{f_1,\ldots,f_k\} : f_i\in \mathcal{F}, i\in\{1,\ldots,k\}\}$, 
	Then for any sample $\widehat D$ of size $n$, we have
		\begin{equation}
	\widehat{\mathfrak{R}}_{\widehat D}(\mathcal G)  \le\sum_{i=1}^{k}\, \widehat{\mathfrak{R}}_{\widehat D}(\mathcal F_i) 
	\end{equation}
\end{lemma}

\begin{theorem}[Lemma 3.6]\label{3.4}
	Let $\mathcal{F} \in \mathbb{R}^{\mathcal X \times \mathcal Y}$ is a hypothesis set. Let $\mathcal{H}$ be the set of classifiers (mapping $\mathcal X$ to $\mathcal Y$) corresponding to $\mathcal{F}$. 
	For any $\delta > 0$, with probability $1-2\delta$, the following holds simultaneously for any scoring function \(f\),
	\begin{equation}
		\begin{aligned}
			 & |d_{f,\mathcal F}^{(\rho)}(\widehat P,\widehat Q) -d_{f,\mathcal F}^{(\rho)}(P,Q)|
			\\
			 \leq& \frac{k}{\rho}{\mathfrak{R}}_{n, P}(\Pi_{\mathcal H}\mathcal F) +
			\frac{k}{\rho}{\mathfrak{R}}_{m, Q}(\Pi_{\mathcal H}\mathcal F)
			+\sqrt{\frac{\log \frac{2}{\delta}}{2n}} +\sqrt{\frac{\log \frac{2}{\delta}}{2m}}.
		\end{aligned}
	\end{equation}
\end{theorem}

\begin{proof}
For $\forall f,f' \in \mathcal{F}$, define the $\tau_f$-transform of $f'$ to be
\[
\tau_f f'(x,y)= \begin{cases}
	 f'(x,1)  \quad & \textnormal{if} \, y = h_f(x) 
	 \\ 
	 f'(x,h_f(x) )  \quad & \textnormal{if} \, y = 1
	 \\ 
	 f'(x,y) \quad & \textnormal{else} \,  
 \end{cases}
\]
where $h_f$ is the induced classifier mapping from $\mathcal{X}$ to $\mathcal{Y}$.
Let $\mathcal{G} = \{ \tau_f f' | f, f'\in\mathcal{F} \}$,  
$\tilde{\mathcal{G}} = \{ (x, y) \mapsto \rho_g(x,y) | g\in\mathcal{G} \}$.
Consider the family of functions $\Phi_{\rho} \comp \tilde{\mathcal{G}} $ which takes values in $[0,1]$.
By Lemma \ref{3.1}, with probability at least $1-\delta$, for $\forall g\in \mathcal{G}$.
\[\begin{aligned}
&|\err_P^{(\rho)} (g) - \err_{\widehat P}^{(\rho)} (g) |
\\
=& |\ex \Phi_{\rho} \comp \rho_{g}(x,y) - 
\frac{1}{n} \sum_{i=1}^{n} \Phi_{\rho} \comp  \rho_{g}(x_i,y_i)|
\\
\leq & 2 \mathfrak R_{n, D}(\Phi_{\rho} \comp \tilde{\mathcal{G}} ) + \sqrt{\frac{\log \frac{2}{\delta}}{2n}}
\end{aligned}\]
Regard all the data as from the same class $1$. Define:
\[
\mathfrak R_{n, D}^{0} (\mathcal{G}) = \ex_{(x_i,1), x_i\sim D^n} \widehat{\mathfrak R}_{ \widehat D} (\mathcal{G}) 
\]
Then the above equations becomes
\[\begin{aligned}
&|\disp_P^{(\rho)} (g, 1) - \disp_{\widehat P}^{(\rho)} (g, 1) |
\\
=&  |\ex \Phi_{\rho} \comp \rho_{g}(x,1) - 
\frac{1}{n} \sum_{i=1}^{n} \Phi_{\rho} \comp  \rho_{g}(x_i,1)|
\\
\leq & 2 \mathfrak R_{n, D}^{0} (\Phi_{\rho} \comp \tilde{\mathcal{G}} ) + \sqrt{\frac{\log \frac{2}{\delta}}{2n}}
\end{aligned}\]
For any $f, f'\in \mathcal{F}$, let $g = \tau_f f'$. Then $g\in \mathcal{G}$ and 
\[
\disp_{P}^{(\rho)} (g,1) = \disp_{P}^{(\rho)} (f',f), \,
\disp_{\widehat P}^{(\rho)} (g,1) = \disp_{\widehat P}^{(\rho)} (f',f)
\]
Thus, 
\[\begin{aligned}
&\sup_{f,f'\in \mathcal{F}} |\disp_{P}^{(\rho)} (f',f) - \disp_{\widehat P}^{(\rho)} (f',f)|
\\
\leq& \sup_{g\in \mathcal{G}} |\disp_{P}^{(\rho)} (g,1) - \disp_{\widehat P}^{(\rho)} (g,1)|
\\
\leq&  2 \mathfrak R_{n, D}^{0} (\Phi_{\rho} \comp \tilde{\mathcal{G}} ) + \sqrt{\frac{\log \frac{2}{\delta}}{2n}}
\end{aligned}\]
By Lemma \ref{3.2}, $\mathfrak R_{n, D}^{0} (\Phi_{\rho} \comp \tilde{\mathcal{G}} ) 
\le \frac{1}{\rho} \mathfrak R_{n, D}^{0} (\tilde{\mathcal{G}} ) $
\[\begin{aligned}
&\quad\mathfrak R_{n, D}^{0} (\tilde{\mathcal{G}}) =\frac{1}{n}\ex_{S,\sigma} (\sup_{g\in \mathcal{G}} \sum_{i=1}^{n} \sigma_i \rho_g(x_i, 1))
\\
&=\frac{1}{n}\ex_{S,\sigma} (\sup_{g\in \mathcal{G}} 
\sum_{i=1}^{n} \sigma_i (g(x_i, 1) - \max_{y\ne 1} g(x_i, y))
\\
&=\frac{1}{n}\ex_{S,\sigma} (\sup_{f,f'\in \mathcal{F}} 
\sum_{i=1}^{n} \sigma_i (f'(x_i, h_f(x_i)) - \max_{y\ne h_f(x_i)}\!\!\!\! f'(x_i, y))
\\
&\le \frac{1}{n}\ex_{S,\sigma} \sup_{f\in \mathcal{F}, h\in \mathcal{H}} 
\sum_{i=1}^{n} \sigma_i f(x_i, h(x_i))
\\
& + \frac{1}{n}\ex_{S,\sigma} \sup_{f\in \mathcal{F}, h\in \mathcal{H}} 
\sum_{i=1}^{n} \sigma_i (- \max_{y\ne h(x_i)} f'(x_i, y))
\\
&= \mathfrak R_{n, D} (\Pi_{\mathcal H} \mathcal{F})
+ \frac{1}{n}\ex_{S,\sigma} \sup_{f\in \mathcal{F}, h\in \mathcal{H}} 
\sum_{i=1}^{n} \sigma_i  \max_{y\ne h(x_i)} f(x_i, y)
\end{aligned}\]
Define the permutation 
\[
\xi(i)=\begin{cases}
 i+1 &i=1,\ldots,k-1
\\ 
 1 & i=k
\end{cases}\]
By our assumption of $\mathcal{H}$, we have the result that $\forall h\in \mathcal{H}$, $\xi^j h\in \mathcal{H},\; j=1,2,\ldots,k-1$.
\[\begin{aligned}
 &\frac{1}{n}\ex_{S,\sigma} \sup_{f\in \mathcal{F}, h\in \mathcal{H}} 
\sum_{i=1}^{n} \sigma_i  \max_{y\ne h(x_i)} f(x_i, y) 
\\
=&  \frac{1}{n}\ex_{S,\sigma} \sup_{f\in \mathcal{F}, h\in \mathcal{H}} 
\sum_{i=1}^{n} \sigma_i  \max_{j\in \{1,\ldots, k-1\}} f(x_i, \xi^j h(x_i))
\end{aligned}\]
Let $\Pi_\mathcal{H}\mathcal{F}^{(k-1)} = \{ \max\{f_1, \ldots, f_{k-1}\} | f_i \in \Pi_{\mathcal{H} }\mathcal F, i=1,\ldots,k-1 \}$. Then applying Lemma \ref{3.3}:
\[\begin{aligned}
	&\frac{1}{n}\ex_{S,\sigma} \sup_{f, h} 
	\sum_{i=1}^{n} \sigma_i  \max_{j\in \{1,\ldots, k-1\}} f(x_i, \xi^j h(x_i))
	\\
	=& \frac{1}{n}\ex_{S,\sigma} \sup_{f\in \Pi_\mathcal{H}\mathcal{F}^{(k-1)}} 
	\sum_{i=1}^{n} \sigma_i   f(x_i)
	\\
	\leq& \frac{k-1}{n} \ex_{S,\sigma} \sup_{f\in \Pi_\mathcal{H}\mathcal{F}} 
	\sum_{i=1}^{n} \sigma_i   f(x_i)
\end{aligned}\]
Therefore, we have
\[\begin{aligned}
\mathfrak R_{n, D}^{0} (\tilde{\mathcal{G}}) \le&  \mathfrak R_{n, D} (\Pi_{\mathcal H} \mathcal{F})+\frac{k-1}{n} \ex_{S,\sigma} \sup_{f\in \Pi_\mathcal{H}\mathcal{F}} 
\sum_{i=1}^{n} \sigma_i   f(x_i)\\
 \le & k \mathfrak R_{n, D} (\Pi_{\mathcal H} \mathcal{F}),\end{aligned}
\]
\[\begin{aligned}&\sup_{f,f'\in \mathcal{F}} |\disp_{P}^{(\rho)} (f',f) - \disp_{\widehat P}^{(\rho)} (f',f)|\\
&\quad\le  \frac{2k}{\rho}{\mathfrak{R}}_{n, P}(\Pi_{\mathcal H}\mathcal F) + \sqrt{\frac{\log \frac{2}{\delta}}{2n}}.\end{aligned}\]

Similarly
\[\begin{aligned}&\sup_{f,f'\in \mathcal{F}} |\disp_{Q}^{(\rho)} (f',f) - \disp_{\widehat Q}^{(\rho)} (f',f)|\\
&\quad\le  \frac{2k}{\rho}{\mathfrak{R}}_{m, Q}(\Pi_{\mathcal H}\mathcal F) + \sqrt{\frac{\log \frac{2}{\delta}}{2m}}.\end{aligned}\]

Therefore, we conclude
\[
	\begin{aligned}
		 & \sup_{f\in\mathcal{F}}|d_{f,\mathcal F}^{(\rho)}(\widehat P,\widehat Q) -d_{f,\mathcal F}^{(\rho)}(P,Q)|\\
		 \leq & \sup_{f,f'\in \mathcal{F}} |\disp_{Q}^{(\rho)} (f',f) - \disp_{\widehat Q}^{(\rho)} (f',f)|\\&+\sup_{f,f'\in \mathcal{F}} |\disp_{P}^{(\rho)} (f',f) - \disp_{\widehat P}^{(\rho)} (f',f)|
		\\
		 \leq & \frac{2k}{\rho}{\mathfrak{R}}_{n, P}(\Pi_{\mathcal H}\mathcal F) +
		\frac{2k}{\rho}{\mathfrak{R}}_{m, Q}(\Pi_{\mathcal H}\mathcal F)
		+\sqrt{\frac{\log \frac{2}{\delta}}{2n}} +\sqrt{\frac{\log \frac{2}{\delta}}{2m}}.
	\end{aligned}
\]

\end{proof}

\begin{theorem}[Theorem 3.7]\label{the:bound2}
	For any $\delta > 0$, with probability $1-3\delta$, we have the following uniform generalization bound for all scoring functions \(f\)
	\begin{equation}
	\begin{aligned}
	\err_Q(f)\leq& \err_{\widehat P}^{(\rho)}(f)+d_{f,\mathcal F}^{(\rho)}(\widehat P,\widehat Q) +\lambda
	\\ 
	+&\frac{2k^2}{\rho}{\mathfrak{R}_{n, P}}(\Pi_1\mathcal F) +\frac{2k}{\rho}{\mathfrak{R}_{n, P}}(\Pi_{\mathcal H}\mathcal F)+2\sqrt{\frac{\log \frac 2 {\delta}}{2n}}
	\\ 
	\ +&\frac{2k}{\rho}{\mathfrak{R}_{m, Q}}(\Pi_{\mathcal H}\mathcal F)+\sqrt{\frac{\log \frac{2}{\delta}}{2m}} .
	\end{aligned}
	\end{equation}
\end{theorem}
\begin{proof}
	This is the result of combining Theorem \ref{the:bound1}, Equation (\ref{equation18}) and Theorem~\ref{3.4}.
\end{proof}

\begin{example}[Linear Classifiers]\label{example}
	Let \[S\subseteq \mathcal{X}=\{\mathbf{x}\in\mathbb{R}^s|\|\mathbf{x}\|_2\leq r\}\] be a sample of size \(m\) and suppose 
	\[
	\begin{aligned}
		\mathcal{F}=\big\{& f :\mathcal{X}\times \{\pm 1\}\to \mathbb{R}\;\big{|}\; f(\mathbf{x},y)=\\ &\qquad\qquad\mathrm{sgn}(y)\ \mathbf{w}\cdot \mathbf{x},\; \|\mathbf{w}\|_2\leq \Lambda \big\},\\
		\mathcal{H}=\big\{ & h\;|\;h(\mathbf{x})=\mathrm{sgn}(\mathbf{w}\cdot\mathbf{x}), \; \|\mathbf{w}\|_2\leq \Lambda\}.
	\end{aligned}\]
	Then the empirical Rademacher complexity of \(\Pi_{\mathcal{H}}\mathcal{F}\) can be bounded as follows:
	\[\widehat{\mathfrak{R}}_{S}(\Pi_{\mathcal{H}}\mathcal{F})\leq 2\Lambda r\sqrt{\frac{d\log \frac{\mathrm{e}m}{d}}{m}},\]
	where \(d\) is the VC-dimension of \(\mathcal{H}\).
	If we further suppose 
	\[\min_{x\in S}|\mathbf{w}\cdot\mathbf{x}|=1\wedge \|w\|_2,\]
	then 
	\[\widehat{\mathfrak{R}}_{S}(\Pi_{\mathcal{H}}\mathcal{F})\leq 2\Lambda^2 r^2\sqrt{\frac{\log \mathrm{e}m}{m}}.\]	
\end{example}

To prove this we need two lemmas.

\begin{lemma}[Propostion 6 of \citet{maurer2016vector}]\label{exlem1}
	Let \(\xi_{i}\) be the Rademacher random variables. For any vector \(\mathbf{v} \in \mathbb{R}^{s},\) the following holds:
	\[\|\mathbf{v}\|_{2} \leq \sqrt{2} \underset{\xi_{i} \sim\{ \pm 1\}, i \in\{1,2,\ldots,s\}}{\mathbb{E}}|\langle\boldsymbol{\xi}, \mathbf{v}\rangle|.\]
\end{lemma}

\begin{lemma}[Theorem 4.2 of \citet{thy:mohri2012foundations}]\label{exlem2}
	Let \(S \subseteq\{\mathbf{x} :\|\mathbf{x}\| \leq r\} .\) Then, the \(V C\)-dimension \(d\) of the set of canonical hyperplanes
	\[\big\{x \mapsto \operatorname{sgn}(\mathbf{w} \cdot \mathbf{x}) : \min _{x \in S}|\mathbf{w} \cdot \mathbf{x}|=1 \wedge\|\mathbf{w}\|_2 \leq \Lambda\big\}\] verifies
	\[d \leq r^{2} \Lambda^{2}.\]
\end{lemma}

Now we present the proof of Example \ref{example}.
\begin{proof}
	By the definition of empirical Rademacher complexity and Cauchy-Schwartz inequality
	\[\begin{aligned}
		m \widehat{\mathfrak{R}}_{S}(\Pi_{\mathcal{H}}\mathcal{F})& 
		= \ex_{\boldsymbol{\sigma}}\sup_{f,h}\sum_{i=1}^{m}\sigma_i f(\mathbf{x}_i,h(\mathbf{x}_i))\\
		& = \ex_{\boldsymbol{\sigma}}\sup_{\mathbf{w},h}\sum_{i=1}^m \sigma_ih(\mathbf{x}_i)\langle \mathbf{w},\mathbf{x}_i\rangle\\
		& =\ex_{\boldsymbol{\sigma}}\sup_{\mathbf{w},h}\langle \mathbf{w}, \sum_{i=1}^{m}\sigma_ih(\mathbf{x}_i)\mathbf{x}_i\rangle\\
		&\leq\ex_{\boldsymbol{\sigma}}\sup_{h}\Lambda\|\sum_{i=1}^{m}\sigma_ih(\mathbf{x}_i)\mathbf{x}_i\|_2
	\end{aligned}\]
	Applying Lemma \ref{exlem1}, we get
	\[\begin{aligned}
		& \ex_{\boldsymbol{\sigma}}\sup_{h}\Lambda \|\sum_{i=1}^m h(\mathbf{x}_i)\sigma_i\mathbf{x}_i\|_2\\
		\leq & \sqrt{2}\Lambda\ex_{\boldsymbol{\sigma}}\sup_{h}\ex_{\boldsymbol{\xi}\sim\{\pm 1\}^s}|\langle \boldsymbol{\xi},\sum_{i=1}^m h(\mathbf{x_i})\sigma_i\boldsymbol{x_i}\rangle |\\
		\leq & \sqrt{2}\Lambda\ex_{\boldsymbol{\sigma}} \ex_{\boldsymbol{\xi}}\sup_{h}|\langle \boldsymbol{\xi},\sum_{i=1}^m h(\mathbf{x}_i)\sigma_i\mathbf{x}_i\rangle|\\
		= & \sqrt{2}\Lambda\ex_{\boldsymbol{\sigma},\boldsymbol{\xi}}\sup_{h}\langle \boldsymbol{\xi},\sum_{i=1}^m h(\mathbf{x}_i)\sigma_i\mathbf{x}_i\rangle\\
		=& \sqrt{2}\Lambda\ex_{\boldsymbol{\sigma},\boldsymbol{\xi}}\sup_{h}\sum_{i=1}^m\sum_{j=1}^s\xi_j\sigma_ih(\mathbf{x}_i)x_{ij}.
	\end{aligned}\]
	Let 
	\[A=\left\{ (h(\mathbf{x}_1),\ldots,h(\mathbf{x}_m))\;|\;h\in \mathcal{H}\right\}.\]
	By Jensen's inequality, for any \(t>0\)
	\[
		\begin{aligned}
			& \exp ( t \ex_{\boldsymbol{\sigma},\boldsymbol{\xi}}\sup_{h}\sum_{i=1}^m\sum_{j=1}^s\xi_j\sigma_ih(\mathbf{x}_i)x_{ij})\\
			\leq & \ex_{\boldsymbol{\sigma},\boldsymbol{\xi}} \exp(t\sup_{h}\sum_{i=1}^m\sum_{j=1}^s\xi_j\sigma_ih(\mathbf{x}_i)x_{ij})\\
			\leq & \ex_{\boldsymbol{\sigma},\boldsymbol{\xi}} \sum_{\mathbf{a}\in A}\exp( t\sum_{i=1}^m\sum_{j=1}^s\xi_j\sigma_i a_i x_{ij} )\\
			=&\sum_{\mathbf{a}\in A}\ex_{\boldsymbol{\sigma},\boldsymbol{\xi}}\prod_{i=1 }^m\prod_{j=1}^s \exp(t\xi_j\sigma_i a_i x_{ij})\\
			=&\sum_{\mathbf{a}\in A}\prod_{i=1 }^m\prod_{j=1}^s \ex_{{\sigma_i},{\xi_j}}\exp(t\xi_j\sigma_i a_i x_{ij})\\
			\leq & \sum_{\mathbf{a}\in A}\prod_{i=1}^m\prod_{j=1}^s\exp( \frac{t^2(a_i x_{ij})^2}{2} )\\
			= &\sum_{\mathbf{a}\in A}\exp( \frac{t^2 \sum_{i=1}^{m}\|\mathbf{x}_i\|_2^2}{2} )\leq |A|\exp( \frac{t^2r^2m}{2} ).
		\end{aligned}\]
	Thus 
		\[\sup_{h}\ex_{\boldsymbol{\sigma},\boldsymbol{\xi}}\sum_{i=1}^m\sum_{j=1}^s\xi_j\sigma_ih(\mathbf{x}_i)x_{ij}\leq \frac{\log |A|}{t}+\frac{tr^2m}{2}.\]
	Take 
	\[t=\sqrt{\frac{2\log|A|}{r^2m}},\]
	We get
	\[\sup_{h}\ex_{\boldsymbol{\sigma},\boldsymbol{\xi}}\sum_{i=1}^m\sum_{j=1}^s\xi_j\sigma_ih(\mathbf{x}_i)x_{ij}\leq \sqrt{2r^2m\log|A|}.\]
	Note that 
	\[\log |A|=\log (\Pi(m,\mathcal{H}))\leq d\log \frac{\mathrm{e}m}{d},\]
	We conclude
	\[\widehat{\mathfrak{R}}_{S}(\Pi_{\mathcal{H}}\mathcal{F})\leq 2\Lambda r\sqrt{\frac{d\log \frac{\mathrm{e}m}{d}}{m}}.\]
	Now if the extra condition is satisfied, by Lemma \ref{exlem2}
	\[\widehat{\mathfrak{R}}_{S}(\Pi_{\mathcal{H}}\mathcal{F})\leq 2\Lambda^2 r^2\sqrt{\frac{\log \mathrm{e}m}{m}}.\]

\end{proof}

\begin{definition}[{Covering Number}]
	Let $(M, d)$ be a metric space. A subset \(\widehat{T}\subseteq M\) is called an \(\epsilon\) cover of \(T\subseteq M\) if for every \(t\in T\), there exists an \(t'\in \widehat T\) such that \(\rho(t,t')\leq \epsilon\). The {covering number} of \(T\) is the cardinality of the smallest \(\epsilon\) cover of \(T\), that is 
	\begin{equation}
	\mathcal N(\epsilon, T,d)\triangleq\min\left\{|\widehat T|\;\Big{|}\;\widehat T \textrm{ is an } \epsilon\textrm{ cover of }T\right\}.
	\end{equation}
	Let \((\mathcal F_{x_1,\ldots,x_n},\mathcal L_2(\widehat D))\) stand for the data-dependent \(\mathcal L_2\) metric space given by metric
	\begin{equation}
	d(f,f')\triangleq\|f-f'\|_2=\sqrt{\frac 1 n \sum_{i=1}^n(f(x_i)-f'(x_i))^2}
	\end{equation}
	where \(x_1,\!\ldots,\!x_n\) are a sample from space \(\mathcal X\) and \(\mathcal F_{x_1,\ldots,x_n}\) stands for the restriction of (real-valued) function class \(\mathcal F\) to that sample.
	Denote the {\(\mathcal L_2\) covering number} by
	\begin{equation}\mathcal N_2(\epsilon,\mathcal F)\triangleq\mathcal N(\epsilon,\mathcal F,\mathcal L_2(\widehat D) ).
	\end{equation}
\end{definition}

The covering number can be interpreted as a measure of the richness of the class \(\mathcal F\) at the scale \(\epsilon\). For a fixed value of $\epsilon$, this covering number, and in particular how rapidly it grows with \(n\), indicate how much the set \(\mathcal F_{x_1,\ldots,x_n}\) ``fills up'' \(\mathbb R^n\), when we examine it at the scale \(\epsilon\).
	
First we show that the \(\mathcal L_2\) covering number of \(\Pi_{\mathcal H}\mathcal F\) can be bounded by that of \(\Pi_1 \mathcal F\) and \(\Pi_1\mathcal H\).

\begin{lemma}\label{3.5}
Suppose the value of \(f\in \Pi_1\mathcal F\) is bounded by \(L<\infty\), i.e.
\begin{equation}
\|f\|_2\leq L.
\end{equation}
Then we have
\begin{equation}\mathcal N_2(\epsilon,\Pi_{\mathcal H}\mathcal F)\leq \mathcal N_2^k(\frac {\epsilon}{2k},\Pi_1 \mathcal F)\cdot\mathcal N_2^k(\frac {\epsilon}{2kL},\Pi_1 \mathcal H)\end{equation}
\end{lemma}
\begin{proof} For any \(g\in \Pi_{\mathcal H}\mathcal F\), \(g=\langle h, f\rangle\), choose \(\widehat h_i, \widehat f_i\) from the \(\mathcal N_2^k(\frac {\epsilon}{2kL},\Pi_1 \mathcal H)\) and \(\mathcal N_2(\frac {\epsilon}{2k},\Pi_1 \mathcal F)\) cover of \(\Pi_1\mathcal H\) and \(\Pi_1\mathcal F\) according to the components \(h_i,f_i\) of \(h,f\) (\(i=1,\ldots k\)). Let \(\widehat g=\sum_{i=1}^k \widehat h_i \widehat f_i\). Then the choices of \(\hat g\) is at most \(\mathcal N_2^k(\frac {\epsilon}{2k},\Pi_1 \mathcal F)\cdot\mathcal N_2^k(\frac {\epsilon}{2kL},\Pi_1 \mathcal H)\). By Minkowski inequality and H\"older inequality we have
\[\begin{aligned}
\|g-\widehat g\|_2&=\|\sum_{i=1}^k(h_if_i-\widehat h_i\widehat f_i)\|_2\\
&=\|\sum_{i=1}^k(h_i(f_i-\widehat f_i)+\widehat f_i(h_i-\widehat h_i)\|_2\\
&\leq \sum_{i=1}^k (\|h_i\|_2\ \|f_i-\widehat f_i\|_2+\|\widehat f_i\|_2\ \|h_i-\widehat h_i\|_2)\\
&\leq \sum_{i=1}^k (\|f_i-\widehat f_i\|_2+L\|h_i-\widehat h_i\|_2)\leq\epsilon.
\end{aligned}\]
\end{proof}

\begin{lemma}[Dudley's Entropy Bound, \citet{thy:talagrand2014upper}]\label{3.7}
 For any function class \(\mathcal F\) containing functions \(f:\mathcal X\to \mathbb R\), we have
\begin{equation}
	\mathfrak R_{n, D}(\mathcal F)\leq \inf_{\epsilon\geq 0}\Big \{4\epsilon +\frac{12}{\sqrt n}\int_{\epsilon}^{\sup_{f\in\mathcal F}\|f\|_2}\!\!\!\!\!\!\!\sqrt{\log \mathcal N_2(\tau,\mathcal F)}\mathrm d \tau\Big\}
\end{equation}
\end{lemma}

\begin{theorem}[Theorem 3.8]\label{3.8}
With the same conditions in Theorem \ref{the:bound2}, further suppose \(\Pi_1\mathcal F \) is bounded in \(\mathcal L_2\) by \(L\). For \(\delta>0\), with probability \(1-3\delta\), we have the following uniform generalization bound for all scoring functions \(f\),
\begin{equation}
	\begin{aligned}
		\err_Q&(f)\leq \err_{\widehat P}^{(\rho)}(f)+d_{f,\mathcal F}^{(\rho)}(\widehat P,\widehat Q)+\lambda+2\sqrt{\frac{\log \frac 2 \delta}{2n}}\\
		&+\sqrt{\frac{\log \frac 2 \delta}{2m}}
		 +\frac{16k^2\sqrt {k}}{\rho}\inf_{\epsilon\geq 0}\Big \{\epsilon+3\big(\frac 1 {\!\sqrt n}\!\!+\!\!\frac 1 {\!\sqrt m}\big)\\
		&\big(\!\!\int_{\epsilon}^L\!\!\!\!\!\!\! \sqrt{\log \mathcal N_2(\tau, \Pi_1\mathcal F)}\mathrm d \tau\!+\!L\!\!\int_{\epsilon/L}^1\!\!\!\!\!\!\! \sqrt{\log \mathcal N_2(\tau,\Pi_1\mathcal H)}\mathrm d \tau\big)\Big\}
	\end{aligned}
\end{equation}
\end{theorem}
\begin{proof}
Directly combine Theorem \ref{3.4}, Lemma \ref{3.5} and Lemma \ref{3.7} and put together similar terms noticing that
\[\sqrt{a+b}\leq \sqrt a +\sqrt b,\]
and the change of variable
\[\int_a^b f(x)\mathrm d x=\int_{a/t}^{b/t} tf(tx)dx.\]
\end{proof}

\begin{definition}[Fat-Shattering Dimension]
We say that \(\mathcal F\) shatters \(x_1,\ldots,x_n\) at scale \(\gamma\), if there exists witness \(s_1,\ldots,s_n\) such that, for every \(\epsilon\in\{\pm 1\}^n\), there exists \(f_{\epsilon}\in \mathcal F\) such that \(\forall t\in\{1,\ldots,n\}\)
\begin{equation}\epsilon_t\cdot (f_\epsilon(x_t)-s_t)\geq \gamma/2.\end{equation}
Then define the fat-shattering dimension
\begin{equation}\begin{aligned}\mathrm{Fat}_\gamma(\mathcal F)\triangleq \max \{n\big{|}&\exists\  x_1,\ldots,x_n\in\mathcal X \;s.t.\\ &\mathcal F\;\gamma\textrm{-shatters } x_1,\ldots,x_n \}.\end{aligned}\end{equation}
\end{definition}

\begin{lemma}\label{3.10}
For any \(\mathcal F\subseteq [-1,1]^{\mathcal X}\) and any \(\gamma\in(0,1)\)
\begin{equation}
	\mathcal N_2(\gamma,\mathcal F)\leq \left(\frac 2 \gamma\right)^{K\ \mathrm{Fat}_{c\gamma}(\mathcal F)}
\end{equation}
where in the above \(c\) and \(K\) are universal constants.
\end{lemma}

\begin{corollary}\label{3.11}
For any \(\mathcal F\subseteq \{-1,1\}^{\mathcal X}\), 
\begin{equation}
	\mathrm{Fat}_{c\gamma}(\mathcal F)=\mathrm{Fat}_0(\mathcal F)=\mathrm{VC}(\mathcal F).
\end{equation}
\end{corollary}

For detailed proofs of these two propositions, refer to \citet{thy:mendelson2003entropy,thy:rakhlin2014statistical}.

\begin{theorem}
Let \(\mathrm{Fat}_{\gamma}(\Pi_1\mathcal F)\) be the fat-shattering dimension of \(\Pi_1\mathcal F\) with scale \(\gamma\) and \(\mathrm{VC}(\Pi_1\mathcal H)\) be the VC-dimension of \(\Pi_1\mathcal H\). Then there exist constants \(C_1,C_2,C_3>0,0<c<1\) independent of \(n,m\) and \(\mathcal F,\mathcal H\) such that for \(\delta>0\), with probability \(1-3\delta\),
\begin{equation}
	\begin{aligned}
	&\err_Q(f)\leq \err_{\widehat P}^{(\rho)}(f)+d_{f,\mathcal F}^{(\rho)}(\widehat P,\widehat Q)+\lambda\\
	&
	+\frac{k^2\sqrt {k}}{\rho}C_1L\big(\frac 1 {\sqrt n}+\frac 1 {\sqrt m}\big)\sqrt{ \mathrm{VC}(\Pi_1\mathcal H)}\\
	&\!\!+\!\!\frac{k^2\!\sqrt {k}}{\rho}\!\!\inf_{\epsilon\geq 0}\!\Big \{\!C_2\epsilon\!+\!C_3\big(\frac 1 {\!\sqrt n}\!\!+\!\!\frac 1 {\!\sqrt m}\!\big)\big(\!\!\!\int_{\epsilon}^L\!\!\!\!\!\sqrt{\mathrm{Fat}_{c\tau}(\Pi_1\mathcal F)\!\log \frac 2 {\tau}}\mathrm d \tau\!\big)\!\Big\}\\
	& +2\sqrt{\frac{\log \frac 2 \delta}{2n}}+\sqrt{\frac{\log \frac 2 \delta}{2m}}.
	\end{aligned}
\end{equation}
\end{theorem}

This results from a direct computation after putting Lemma \ref{3.10} and Corollary \ref{3.11} into Theorem \ref{3.8}.

\section{Analysis of Algorithm}
In this section, we show that \(\gamma\) controls the margin \(\rho\) and that the minimization of this loss will lead to consistency between source and target domain using similar methods with \citet{cite:NIPS14GAN}.

\begin{proposition}
Consider the optimization problem we have defined 
\begin{equation}\label{loss:gamma}
	\max_{f'} \; \gamma\ex_{\widehat{P}}\log (\sigma_{h_f}\comp f')+\ex_{\widehat{Q}}\log (1-\sigma_{h_f}\comp f').
	\end{equation}
Assume that there is no restriction for the choice of $f'$ and $\gamma > 0$. Fixing a single-output classifier $h_{f}$, we have the following two results:
\begin{itemize}
\item[1.] The optimal value of $\sigma_{h_f}\comp f'$ on data $x$ is 
\begin{equation}\label{theory: equilibrium}
\frac{\gamma p(x)}{\gamma p(x) + q(x)}
\end{equation}
where \(p(x)\) and \(q(x)\) are the density functions.
\item[2.] Problem (\ref{loss:gamma}) is equivalent to a $\gamma$-balanced Jensen-Shannon Divergence:
\begin{equation}
\gamma \mathrm {\mathrm{KL}}\Big(P\Big{\|} \frac{\gamma P + Q}{\gamma + 1}\Big) +\mathrm {\mathrm{KL}}\Big(Q\Big{\|} \frac{\gamma P + Q}{\gamma + 1}\Big)
\end{equation}
and has the global minimum at $P=Q$.
\end{itemize}
\end{proposition}
\begin{proof}
\[\begin{aligned}
    &\gamma \ex_{P}\log (\sigma_{h_f}\comp f')+\ex_{Q}\log (1-\sigma_{h_f}\comp f')
    \\
    &=\int_{x \in \mathcal{X}} \gamma p(x) \log (\sigma_{h_f}\comp f') +q(x)\log (1-\sigma_{h_f}\comp f')\ \mathrm d x
    \\
    & = \int_{x \in \mathcal{X}} \mathrm J(x, \sigma_{h_f}\comp f')\  \mathrm d x.
\end{aligned}\]
For there is no restriction on $f'$, we could find the $f'$ that reaches the maximum on each $x\in \mathcal{X}$. Simple calculus gives the result that for $\forall x \in \mathcal{X}$, $\mathrm J(x, \sigma_{h_f}\comp f')$ is the largest when 
\[\sigma_{h_f}\comp f' =  \frac{\gamma p(x)}{\gamma p(x) + q(x)}.\]
So the first conclusion is proved. At this time, 
\[\begin{aligned}
&\mathrm J(x, \sigma_{h_f}\comp f') 
\\
=&  \gamma p(x) \log (\sigma_{h_f}\comp f') +q(x)\log (1-\sigma_{h_f}\comp f')
\\
=&\gamma p(x) \log (\frac{\gamma p(x)}{\gamma p(x) + q(x)}) +q(x)\log (\frac{ q(x)}{\gamma p(x) + q(x)})
\\
=&\gamma p(x) \log (\frac{ p(x)}{\frac{\gamma p(x) + q(x)}{\gamma  + 1}}) +q(x)\log (\frac{ q(x)}{\frac{\gamma p(x) + q(x)}{\gamma  + 1}}) 
\\
& + \gamma\log \gamma \, p(x) - (\gamma p(x) +q(x)) \log (\gamma + 1).
\end{aligned}\]
Notice that $\frac{\gamma p(x) + q(x)}{\gamma  + 1}$ is density of mixed distribution
\[
\frac{\gamma P + Q}{\gamma  + 1}.
\]
Integrate $\mathrm J(x, \sigma_{h_f}\comp f') $ on the $\mathcal{X}$,
\[\begin{aligned}
&\int_{x \in \mathcal{X}} \mathrm J(x, \sigma_{h_f}\comp f')\ \mathrm d x
\\
	=&\int_{x \in \mathcal{X}} [\gamma p(x) \log (\frac{ p(x)}{\frac{\gamma p(x) + q(x)}{\gamma  + 1}}) +q(x)\log (\frac{ q(x)}{\frac{\gamma p(x) + q(x)}{\gamma  + 1}})\ \mathrm d x 
	\\
	 &+ \gamma\log \gamma \, p(x) - (\gamma p(x) +q(x)) \log (\gamma + 1)]
	\\
	= &\gamma \mathrm{KL}(P \| \frac{\gamma P + Q}{\gamma  + 1}) + \mathrm{KL}(Q \| \frac{\gamma P + Q}{\gamma  + 1}) 
	\\
	 &+  \gamma\log \gamma - (\gamma + 1) \log (\gamma + 1)]
	 \\
	 	= &\gamma \mathrm{KL}(P \| \frac{\gamma P + Q}{\gamma  + 1}) + \mathrm{KL}(Q \| \frac{\gamma P + Q}{\gamma  + 1}) + C(\gamma),
\end{aligned}\]
where $C(\gamma )$ is a constant only depending on $\gamma$. This derivation shows that the second conclusion holds. 
\end{proof}

This proposition implies that different choices of \(\gamma\) does not lead to mismatch between $P$ and $Q$.

Next we show that \(\gamma\) decides the margin \(\rho\) at equilibrium by analyzing the training process, which ensures the optimality of the margin disparity discrepancy that we achieved after training.

During the training session, the discrepancy between source and target features decreases and converges to a value close to zero, indicating $\psi(P)\approx \psi(Q)$. If $\gamma = 1$, the value of $\sigma_{h_f}\comp f'$ converges to a number around $\frac{1}{2}$ on both the source and target domains, in which case the output of $f'$ for the class predicted by $f$ is probably the largest among all classes. However, the margin of $f'$ might still be close to zero as there might exist a prediction for another class approaching $\frac{1}{2}$ from below. As a result, the value of the margin disparity discrepancy measured by $f'$ does not reach minimization for any \(\rho\).
For $\gamma > 1$, after some calculation, we conclude that the value of $\sigma_{h_f}\comp f' $ will reach $\frac{\gamma}{\gamma +1}$ and the margin of $f'$ will be around $\log \gamma$ at equilibrium as shown in the proposition below.

\begin{proposition}
For any \(j\in\{1,2,\ldots,k\}\) if $\sigma_{j}\comp f > \mu > \frac{1}{2}$, then $f$ is a classifier with margin $\log\frac{\mu}{1-\mu}$.
\end{proposition}

\begin{proof}
For any $r \ne j$, $r\in {1,\ldots,k}$
\[\begin{aligned}
	\mu &< \frac{\sigma_{j} }{\sum_{i=1}^{k} \sigma_i}\\
	&\le \frac{\sigma_{j} }{\sigma_{j} + \sigma_{r}} \\
	&= \frac{1}{1 + e^{ f(x,r) - f(x,j) } }.
\end{aligned}\]
Thus
\[\begin{aligned}
	f(x,j) - f(x,r) > \log (\frac{\mu}{1-\mu} ).
\end{aligned}\]
\end{proof}

\section{Additional Experiments}
We also test our algorithm by minimizing the original MDD loss. Since gradient saturation using the margin losses is fatal in the early training stage, we implement by switching to the margin losses after 2000 steps. The results on Office-31 with the margin \(\rho=\log 4\) (equivalent to \(\gamma=4\)) are reported in Table \ref{table:gamma}. 

\begin{table}[H]
	\centering
	\addtolength{\tabcolsep}{5pt}
	\caption{Accuracy (\%) on {Office-31} with original MDD loss.}
	\label{table:gamma}
	\vskip 0.05in
	\begin{tabular}{cccc}
		\toprule
		 Task  & Accuracy \\
		\midrule
		A $\to$ W & 94.1 \\
		A $\to$ D & 91.8 \\
		D $\to$ W & 100 \\
		W $\to$ D & 98.2 \\
		D $\to$ A & 73.7 \\
		W $\to$ A & 71.7 \\
		Average & 88.3 \\ 
		\bottomrule
	\end{tabular}
\end{table}

\bibliography{paper}
\bibliographystyle{icml2019}